\documentclass{article} % For LaTeX2e
\usepackage{collas2025_conference,times}
\usepackage{easyReview}

\usepackage{amsmath,amsfonts,bm}

\def\eqref#1{equation~\ref{#1}}
\def\1{\bm{1}}

\def\vs{{\bm{s}}}

\def\mD{{\bm{D}}}

\def\mX{{\bm{X}}}

\DeclareMathAlphabet{\mathsfit}{\encodingdefault}{\sfdefault}{m}{sl}
\SetMathAlphabet{\mathsfit}{bold}{\encodingdefault}{\sfdefault}{bx}{n}

\usepackage{amsmath}

  \def\mD{{\mathcal D}}

  \def\mX{{\mathcal X}}

  \DeclareMathAlphabet\mathbfcal{OMS}{cmsy}{b}{n}
  \def\0{{\bf 0}}
  \def\1{{\bf 1}}

  \def\bx{{\bf x}}

  \def\bx{{\bf x}}

\def\eg{\emph{e.g.}} \def\Eg{\emph{E.g.}}
\def\ie{\emph{i.e.}} \def\Ie{\emph{I.e.}}
\def\cf{\emph{c.f.}} \def\Cf{\emph{C.f.}}
\def\etc{\emph{etc.}} \def\vs{\emph{vs.}}
\def\wrt{{w.r.t.~}} \def\dof{d.o.f}
\def\etal{\emph{et al.}}

\usepackage{tikz}
\usepackage{algorithm}
\usepackage{algorithmic}

\usepackage{graphicx}
\usepackage{amsmath}
\usepackage{amssymb}
\usepackage{booktabs}

\usepackage{mathtools}
\usepackage{bm}

\usepackage{balance}
\usepackage{xspace}
\usepackage{pifont}

\usepackage{xspace}
\usepackage{pifont}

\usepackage{amsmath}
\usepackage{subcaption}

\usepackage{threeparttable}
\usepackage{multirow}
\usepackage{booktabs} % for professional tables
\usepackage{bbding}
\usepackage{threeparttable}
\usepackage{rotating}
\usepackage{xcolor,colortbl}
\definecolor{Gray}{gray}{0.85}
\definecolor{LightCyan}{rgb}{0.88,1,1}

\usepackage{makecell}

\newcommand\imad[1]{\textcolor{black}{#1}}

\newlength\myindent
\usepackage{hyperref}
\hypersetup{
    colorlinks=true,
    linkcolor=red,
    filecolor=magenta,
    urlcolor=blue,
    citecolor=purple,
    pdftitle={Overleaf Example},
    pdfpagemode=FullScreen,
    }

\title{Enhancing Plasticity for First Session Adaptation Continual Learning}

\author{Imad Eddine Marouf$^{1}$ \\
\And % Use And to have authors side by side
Subhankar Roy$^{2}$  \\
\And % Use AND to have authors block one under the other
Stéphane Lathuilière$^{1,3}$ \\
\And
Enzo Tartaglione$^{1}$ \\
\AND
 \\
$^{1}$LTCI, Télécom-Paris, Institut Polytechnique de Paris, France\\
$^{2}$University of Bergamo, Italy \\
$^{3}$Inria at University Grenoble Alpes, LJK, France \\
}

\collasfinalcopy % Uncomment for camera-ready version, but NOT for submission.

\begin{document}

\newcommand{\method}{PLASTIC\xspace}
\newcommand{\adam}{ADAM\xspace}
\newcommand{\methodname}{Plasticity-Enhanced Test-Time Adaptation in Class-Incremental Learning\xspace}

\newcommand{\name}{{\sc PLASTIC }}
\newcommand{\mame}{{\sc PLASTIC}}

\newcommand*\circled[1]{\tikz[baseline=(char.base)]{\node[fill=blue!30,shape=circle,draw,inner sep=0.5pt] (char) {#1};}}

\newcommand{\aug}{a}
\newcommand{\augset}{\mathcal{T}}
\newcommand{\inspace}{\mathcal{X}}
\newcommand{\model}{f}
\newcommand{\outspace}{\mathcal{Y}}
\newcommand{\params}{\theta}
\newcommand{\paramsspace}{\Theta}
\newcommand{\U}{\mathcal{U}}

\newcommand{\prepr}[1]{}

\newcommand{\x}{{\bf x}}
\newcommand{\z}{{\bm z}}
\newcommand{\m}{{\bm m}}
\newcommand{\aaa}{{\bm a}}
\newcommand{\w}{{\bm w}}
\newcommand{\p}{{\bm p}}
\newcommand{\cc}{{\bf c}}
\newcommand{\uu}{{\bf u}}
\newcommand{\y}{{\bf y}}
\newcommand{\X}{{\bf X}}
\newcommand{\D}{\mathcal{D}}
\newcommand{\Y}{\mathcal{Y}}
\newcommand{\Sup}{\mathcal{S}}
\newcommand{\Q}{\mathcal{Q}}
\newtheorem{example}{Example}
\newtheorem{theorem}{Theorem}
\newcommand{\firstcite}[1]{\cite{#1}}
\newcommand{\bfname}[1]{{\bf #1}}

\def\eg{\emph{e.g.}} \def\Eg{\emph{E.g.}}
\def\ie{\emph{i.e.}} \def\Ie{\emph{I.e.}}
\def\cf{\emph{c.f.}} \def\Cf{\emph{C.f.}}
\def\etc{\emph{etc.}} \def\vs{\emph{vs.}}
\def\wrt{{w.r.t.~}} \def\dof{d.o.f}
\def\etal{\emph{et al.}}

\maketitle
\begin{abstract}
The integration of large pre-trained models (PTMs) into Class-Incremental Learning (CIL) has facilitated the development of computationally efficient strategies such as First-Session Adaptation (FSA), which fine-tunes the model solely on the first task while keeping it frozen for subsequent tasks. Although effective in homogeneous task sequences, these approaches struggle when faced with the heterogeneity of real-world task distributions. We introduce Plasticity-Enhanced Test-Time Adaptation in Class-Incremental Learning (\method), a method that reinstates plasticity in CIL while preserving model stability. \method leverages Test-Time Adaptation (TTA) by dynamically fine-tuning LayerNorm parameters on unlabeled test data, enabling adaptability to evolving tasks and improving robustness against data corruption. To prevent TTA-induced model divergence and maintain stable learning across tasks, we introduce a teacher-student distillation framework, ensuring that adaptation remains controlled and generalizable. Extensive experiments across multiple benchmarks demonstrate that \method consistently outperforms both conventional and state-of-the-art PTM-based CIL approaches, while also exhibiting inherent robustness to data corruptions. Code is available at: \href{https://github.com/IemProg/PLASTIC}{https://github.com/IemProg/PLASTIC}.
\end{abstract}

\section{Introduction}
\label{intro}

In real-world applications, neural networks must continuously adapt to dynamic data streams that introduce new classes over time~\citep{gomes2017survey}. However, sequentially updating a model to accommodate new classes often leads to \textit{catastrophic forgetting} (CF)—a phenomenon where newly acquired knowledge overwrites and erases previously learned information~\citep{french1999catastrophic}. To address this challenge, Class-Incremental Learning (CIL) has been introduced, enabling models to learn new classes continually from evolving data without losing the ability to classify previous classes using a unified classifier~\citep{masana2022class}.

\begin{figure}[H]
    \centering
    \begin{minipage}{0.48\textwidth}
    Given the inevitability of evolving task distributions, this work aims \textit{to address the plasticity limitation of FSA approaches} when dealing with heterogeneous task distributions. To this end, we propose a novel approach that preserves the stability of FSA while introducing the necessary plasticity for adapting to diverse tasks. Our method, Plasticity-Enhanced Test-Time Adaptation in Class-Incremental Learning (\method), leverages Test-Time Adaptation (TTA)~\citep{zhao2023pitfalls,sun2020test,zhang2021memo} to boost model plasticity using only unlabeled data from new tasks. TTA, which was originally designed for mitigating \textit{domain shift}~\citep{quinonero2008dataset} between a pre-trained source model and the unlabelled data in the target domain~\citep{liang2023comprehensive}.
    \vspace{-3mm}
    \end{minipage}
    \hfill
    \begin{minipage}{0.48\textwidth}
    \vspace{-7mm}
        \includegraphics[width=\linewidth]{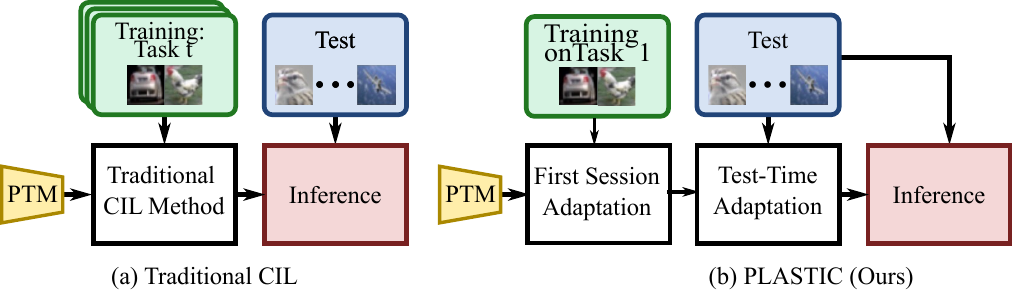}
        \caption{\small Comparison of traditional CIL methods with \method (Ours). (a) Traditional CIL methods entail training on each task's training data, making it prone to forgetting. (b) In \method, we train the PTM only on the first task using Adapters, then adjust them for the subsequent tasks using Test-Time Adaptation and knowledge-distillation for better \textit{stability-plasticity} trade-off.} 
        \label{figure:teaser}
    \end{minipage}%
\end{figure}

Traditional CIL methods often involve training networks from scratch on each new task (see Fig.~\ref{figure:teaser} (a)), employing regularization techniques to mitigate CF~\citep{masana2022class,wang2023comprehensive}. However, the growing adoption of pre-trained models (PTMs) has transformed the CIL landscape, leveraging the strong generalization capabilities of PTMs to enhance learning efficiency. Many PTM-based CIL methods~\citep{wang2022dualprompt,wang2022learning,wang2022s,villa2023pivot,CODAPrompt} freeze the PTM backbone entirely and train only a small set of task-specific learnable parameters to condition the frozen representations. While these methods outperform traditional CIL approaches, they still require task-by-task fine-tuning, which remains computationally expensive, particularly in long task sequences.

To further reduce computational cost in CIL, recent works have shown that using a simple nearest class mean (NCM)~\citep{mensink2013distance} classification with features extracted from PTMs can provide competitive results~\citep{janson2022simple,AdamAdapter}. This approach is further enhanced through \textit{first session adaptation} (FSA) -- often producing state-of-the-art results -- consisting of fine-tuning the PTM \textit{only} on the first task~\citep{AdamAdapter,panos2023first, ranpac} while keeping it frozen for the rest to prevent CF. In essence, FSA aims at (i) \textit{bridging the distribution gap} between the PTM pre-training and downstream tasks by training adapters~\citep{chenadaptformer} on the first task, and (ii) \textit{not risking forgetting} by confining fine-tuning only to the first task and keeping the PTM frozen thereafter. While FSA has proven effective, particularly for \imad{\textit{homogeneous} task sequences—where all tasks share a similar data distribution (e.g., CIFAR-100 or ImageNet-A)—it assumes distributional uniformity across tasks. This assumption breaks down in \textit{heterogeneous} settings, where tasks stem from diverse datasets (e.g., VTAB datasets), resulting in substantial distribution shifts. Consequently, FSA methods, which rely on a frozen PTM after the first task, lack plasticity in such settings and struggle to adapt to evolving task distributions.}
 
TTA proves to be a natural fit for CIL by addressing its dual challenge: maintaining stability while ensuring adaptability to novel tasks. Specifically, \method leverages TTA to achieve:
\textbf{(i)} \textbf{plasticity} by fine-tuning only a small subset of parameters of the PTM on the unlabelled test instance to adapt to any new task at hand; \textbf{(ii)} \textbf{stability} by employing a teacher-student distillation framework, where the teacher model guides adaptation and prevents the student from deviating excessively from the pre-trained representations, ensuring that task-specific updates remain stable while preserving generalization. However, naïve resetting after each TTA step discards past adaptations, preventing meaningful knowledge accumulation across tasks. To address this limitation, \method integrates student-teacher distillation, ensuring that adaptation remains continual rather than isolated to individual tasks. Specifically, the student model, initialized from the adapted PTM, undergoes TTA and is reset after each update. 
This design strikes a critical balance: on one hand, the student-teacher framework retains past learned representations, mitigating TTA’s limitation of discarding adaptation history. On the other hand, reinitializing the student from the PTM at each step preserves its pristine, generalizable representations, preventing harmful drift and ensuring robust adaptation across tasks. The framework is illustrated in Fig.~\ref{figure:teaser} (b). Unlike Continual Test-Time Adaptation (CTTA)~\citep{wang2022continual,song2023ecotta}, which primarily addresses domain shifts by adapting models to evolving environmental conditions, \method is designed for CIL. We extend FSA by integrating TTA and incorporating model resetting mechanisms, thereby enabling the effective integration of new classes while maintaining model stability.

Unlike many CIL methods, \method is intuitive and simple to implement, yet delivers several benefits: \textbf{(i)} it reduces computation during training time by limiting the training to just the first task; \textbf{(ii)} Unlike FSA, \method allows greater adaptability to new tasks without excessive forgetting; \textbf{(iii)} By leveraging TTA, \method inherently withstands corruptions and perturbations~\citep{hendrycks2019benchmarking}, a critical aspect often overlooked in CIL; and \textbf{(iv)} it is highly parameter-efficient.  
Through extensive experiments, we demonstrate that \method achieves state-of-the-art performance across standard CIL benchmarks, surpassing FSA methods. Moreover, by relying on TTA, \method is inherently robust to data corruptions, enhancing its real-world applicability.

\section{Related Work}

\noindent\bfname{Class-Incremental Learning} is an active area of research (see the surveys in~\citep{masana2022class,belouadah2021comprehensive,wang2023comprehensive, parisi2019continual}) that deals with incrementally expanding the knowledge of a model to recognize new classes by training on labeled data arriving in sessions. While learning the new classes the model tends to forget previously acquired information due to the phenomenon called CF~\citep{french1999catastrophic}. Therefore, CIL methods aim at simultaneously balancing the knowledge acquisition (\ie, \textit{plasticity}) and retention (\ie, \textit{stability}) capabilities of the model~\citep{mermillod2013stability}. 
The CIL methods can be broadly categorized into four distinct groups.
(i) Weight regularization CIL methods~\citep{kirkpatrick2017overcoming,liu2018rotate,zenke2017continual,lee2020continual,aljundi2018memory,chaudhry2018riemannian} aim at preventing a drift in the weights of the network, which are relevant to the previous tasks. (ii) Data regularization CIL methods~\citep{jung2016less,li2017learning,rebuffi2017icarl,hou2019learning,castro2018end,liu2020mnemonics,dhar2019learning,douillard2020podnet} instead aim at preventing drift in the network activation by employing knowledge distillation~\citep{hinton2015distilling}. (iii) Memory rehearsal CIL methods reduce forgetting by storing and replaying a small number of exemplars~\citep{rebuffi2017icarl,buzzega2020dark,castro2018end,prabhu2020gdumb}, or by generating synthetic images~\citep{shin2017continual,ostapenko2019learning} or features~\citep{xiang2019incremental}. (iv) Architecture growing CIL methods~\citep{rusu2016progressive,mallya2018packnet,mallya2018piggyback,schwarz2018progress}  dynamically increase the capacity of the network through allocating task-specific learnable parameters to prevent interference between tasks, and hence reduce forgetting. Our proposed \method is unique from most of the existing CIL literature as it operates directly at test-time. Although similar in spirit to GDumb~\citep{prabhu2020gdumb}, which re-trains the network at test-time from scratch using exemplars from the memory, our \method is \textit{rehearsal-free} and requires only a single gradient update on the test sample.

\noindent\bfname{CIL with Large Pre-Trained Models.} With the introduction of large PTMs~\citep{dosovitskiy2020} the field of CIL has seen rapid progress~\citep{wang2022learning,wang2022dualprompt,seale2022coda,AdamAdapter,zhou2022learning,SLCA}. It is mainly driven by the fact that PTMs are highly generalizable to downstream tasks and tuning only a small subset of parameters is sufficient to obtain good performance. Broadly speaking, all the PTM-based CIL methods can be classified under two categories. The \textit{first category} consists of prompt-tuning~\citep{jia2022visual} based CIL methods~\citep{wang2022learning,wang2022dualprompt,seale2022coda, jung2023generating} that keep the PTM frozen and aim to train a set of learnable tokens (or prompts) on the training data to learn task-specific features. At inference, they query the prompt pool to retrieve a prompt pertaining to the test instance, and then condition the frozen PTM with the selected prompt. The prompt-tuning methods mostly differ in how the prompts are learned and how they are selected during inference. For instance, DualPrompt~\citep{wang2022dualprompt} decomposes the prompts into general and expert prompts, whereas CODA-Prompt~\citep{CODAPrompt} refines the prompt selection using an attention mechanism. EASE~\citep{zhou2024expandable} proposes to adaptively learn the projection subspace per task by injecting class-specific adapters in the PTM, to favor class separability-at the cost of more learnable parameters. The \textit{second category} consists of conceptually simpler methods like \adam~\citep{AdamAdapter} and FSA \citep{panos2023first} that fine-tune adapters~\citep{houlsby2019parameter} on the first task and keep the model unaltered for the remainder of tasks. RanPAC~\citep{ranpac} extends this paradigm by training solely on the first task while applying random projections to subsequent tasks. 
CF is prevented by limiting the fine-tuning to only the first task at the cost of reduced plasticity. Differently from the existing methods, \method encourages plasticity via Test-Time Adaptation on every new task. 

\noindent\bfname{Test-Time Adaptation} (TTA) has been proposed to improve the performance of a pre-trained model (trained on \textit{source} data) on out-of-distribution test (or \textit{target}) data, exhibiting domain-shift, by adjusting the model parameters using the unlabelled test samples~\citep{liang2023comprehensive,wang2021tent,sun2020test,liu2021ttt++,zhang2021memo,chen2022contrastive,bartler2022mt3}. In particular, the co-variate shift is simulated with synthetic corruptions~\citep{hendrycks2019benchmarking} (\eg, Gaussian noise, blur, shot noise) or natural shifts (\eg, sim to real~\citep{peng2017visda}). The common theme among all TTA methods is to optimize an unsupervised loss (\eg, entropy minimization~\citep{grandvalet2004semi}) using the test instance and update either all the parameters of the network~\citep{zhang2021memo} or only a subset (\eg, Batch Normalization layers~\citep{wang2021tent}). Post adaptation, the resulting model is used for inference on the test instance, after which it is either reset back to the original checkpoint~\citep{zhang2021memo} or directly used for the next adaptation steps~\citep{wang2021tent}. Different from the original motivation of TTA -- to reduce domain shift between train and test data having overlapping classes -- we exploit the generalizable PTM in CIL to adapt it to tasks (or datasets) containing a non-overlapping set of classes. Moreover, CIL differs from Continual Test-Time Adaptation (CTTA) methods such as CoTTA~\citep{wang2022continual, song2023ecotta}\imad{, \citep{machireddy2022continual}, MetaMix\citep{wang2023metamix} and RMT~\citep{dobler2023robust}}, which address domain shifts under a fixed label space. In contrast, CIL involves an evolving label space with each task. \imad{Additionally, while CTTA methods employ teacher-student frameworks with the teacher model as an exponential moving average (EMA) of the student, our approach utilizes the teacher to actively guide the student's learning through a Kullback-Leibler (KL) divergence loss, enhancing adaptation in CIL scenarios.} In contrast, we focus on adapting closed-vocabulary models (with Adapters~\citep{chenadaptformer}) at TTA using entropy-minimization~\citep{zhang2021memo} and further equip it with student-teacher distillation for exploiting past knowledge.

%Moreover, CIL is also different from Continual TTA~\citep{wang2022continual,song2023ecotta, dobler2023robust}, where the latter considers each task as a different kind of corruption but under the same label space, while in CIL the label space evolves with each task. A recent work~\citep{singh2024controllingforgettingtesttimedata} also leverages TTA for CIL by estimating sparse masks for open-ended models with distillation at test-time. 

%

%To the best of our knowledge, we are the first to demonstrate how TTA can also be seamlessly used for CIL and the effectiveness in CIL.

%Established test-time training methodologies, such as TTT~\citep{sun2020test} and TTT++~\citep{liu2021ttt++}, cohesively train a primary model using both supervised and self-supervised targets. These techniques adapt models exclusively using test data and encompass strategies like batch-norm statistics adaptation~\citep{wan2021efficient, wang2021tent}, test-time entropy minimization~\citep{wang2021tent,fleuret2021test}, ensuring prediction consistency across varied augmentations~\citep{zhang2021memo}. Our research aligns with this full test-time adaptation framework. We propose a new paradigm for CIL using TTA in order to make CIL methods more robust to domain shifts, and our results showed that it improves even their performance on clean data.
\section{Proposed Framework}
\label{sec:ourmethod}
\subsection{Problem Formulation and Overview}

CIL involves learning from a data stream that introduces new classes and aims to construct a unified classifier~\citep{rebuffi2017icarl}. The training process consists of a sequence of $T$ training tasks, denoted as $\D = \{\D^{1}, \D^{2}, \cdots, \D^{T}\}$ with incremental data $\D^{t}$ for each new task, each composed of $K^t$ classes; hence, we have $K=\sum_{t=1}^T K^t$ total number of classes. For each class $k$, the number of training samples is $N_k$. 

In this context, $\x_i$ refers to a training instance belonging to the class $y_i \in Y^t$, where $Y^t$ is the label space of task $t$. In our case, there is no overlap in the label spaces between different tasks, \ie, $Y^t\!\cap\!Y^{t^\prime}\!=\!\varnothing$ for $t\!\neq\!t^\prime$. During the $t$-th training stage, only data from $\D^t$ can be accessed for model updates. Following the approach as in~\citep{wang2022learning,wang2022dualprompt}, we assume the availability of a PTM, such as a ViT~\citep{dosovitskiy2021image} trained on the ImageNet-21K dataset~\citep{ridnik2021imagenet21k}. This PTM serves as the initialization for the CIL model. A successful CIL model $f$ acquires knowledge from new classes while preserving it from the previously encountered ones. The evaluation of the model's capability is performed across all seen classes, denoted as $\mathcal{Y}^T\!=\!Y^1\!\cup\!\cdots\!\cup\!Y^{t-1}\!\cup\!Y^t$, after each task $t$. 

\begin{figure*}[t]
     \centering
     \begin{subfigure}[b]{0.38\textwidth}
         \centering
         \includegraphics[scale=0.57]{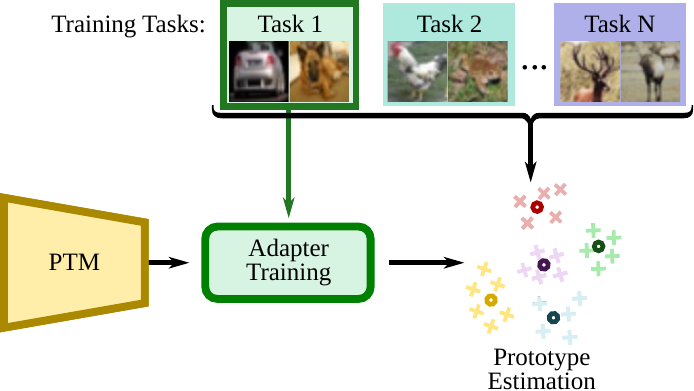}
         \caption{Phase I: Adapter-Based Learning on Task 1}
         \label{fig:y equals x}
     \end{subfigure}
     \hfill
     \begin{subfigure}[b]{0.58\textwidth}
         \centering
         \includegraphics[scale=0.57]{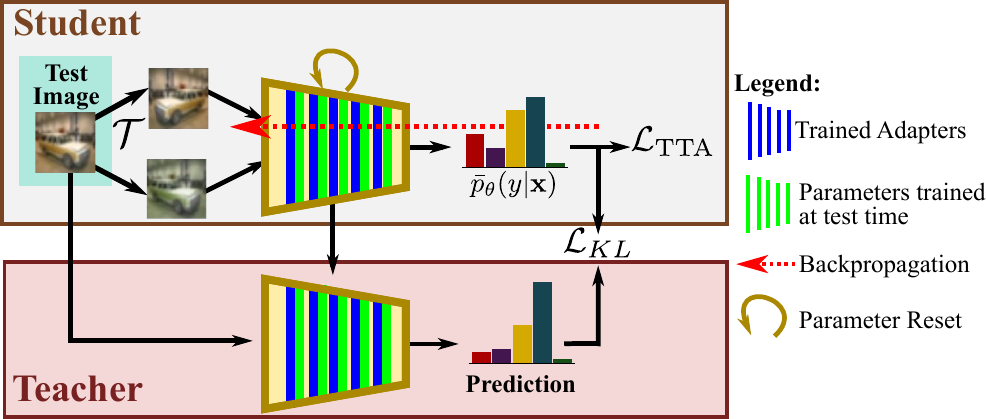}
         \caption{Phase II: Continual TTA with Teacher-Student Learning}
         \label{fig:three sin x}
     \end{subfigure}
        \caption{Our \method is composed of two phases. In \textbf{Phase I}, we fine-tune adapters on the first task's training data in order to learn task-specific features. In \textbf{Phase II}, we perform TTA directly on the unlabelled test instances of each task. This allows the network to adapt to the intricate task-specific features from subsequent tasks, while not deviating too far from the generalizable PTM representations by using a teacher-student distillation framework, where the teacher model ensures stable learning.}
        \label{fig:three graphs}
        \vspace{-2mm}
\end{figure*}

To address the challenge of balancing plasticity and stability in PTMs, we introduce \mame{}, a novel method that employs a two-phase adaptation strategy. In Phase I, we adapt the PTM's feature representations on the first task of the incremental setting (Sec.~\ref{sec:phase1}), as in FSA \citep{panos2023first}. Then, in Phase II we refine the model representations to counteract domain shifts and concept drift (Sec.~\ref{sec:phase2}) using a unique combination of TTA and knowledge distillation. This comprehensive approach, summarized in Fig.~\ref{fig:three graphs} and Algorithm~\ref{alg:overall}, enables our model to maintain good stability, and enhanced plasticity for non-homogeneous tasks.

\subsection{Phase I: Adapting PTMs to New Tasks}
\label{sec:phase1}
%\vspace{-2mm}

While NCM is a strong baseline in CIL \citep{janson2022simple}, it offers no plasticity. The performance is unsatisfactory when downstream datasets exhibit domain shifts, especially with significant concept drift~\citep{alfassy2022feta, hendrycks2021natural}. To allow some plasticity, we employ an adaptation process, as in FSA methods \citep{AdamAdapter,panos2023first}, that bridges the distribution gap between the pre-training and the downstream task distribution. Fine-tuning is limited to the first task as repeated (or sequential) fine-tuning can trigger CF and weaken generalization~\citep{ranpac,kumarfine, AdamAdapter}.

Formally, let $E_{\theta}$ be the PTM model's feature extractor. We first compute class prototypes 
$\boldsymbol{c}_k$ for $k\!=\!1$ by averaging the embeddings of each class in the first task (details in Supp. Mat.). 

We adapt $E_{\theta}$ by inserting lightweight modules inside each layer called Adapters~\citep{adapterNLP}. The adapter parameters, denoted as $\phi$, are learned on the data $\D^1$ while the parameters $\theta$ of $E_{\theta}$ remain frozen. The adaptation is done by minimizing a Cross-Entropy loss using stochastic gradient descent. Importantly, the parameters $\phi$ represent a tiny proportion of the total number of parameters in $E_{\theta,\phi}$ (about $1\%-3\%$) to prevent over-fitting on $\D^1$. This procedure provides an adapted feature extractor $E^{*}_{\theta,\phi}$ with domain-specific knowledge, while keeping the total number of trainable parameters minimal.

\subsection{Phase II: Continual Test-Time Model Refinement}
\label{sec:phase2}
At the end of Phase I, the adapted PTM is adept at classifying samples from the first task and exhibits strong performance on the subsequent tasks with the NCM classifier under the assumption that task distributions remain unaltered. Given that this assumption is vacuous when subsequent tasks are heterogeneous in nature, we propose to improve the plasticity with continual test-time model refinement by directly using the unlabelled test instances. This phase consists of test-time refinement and teacher-student distillation, which are explained below.

\textbf{Test-time refinement.} After the initial adaptation on the first task, our model may still provide unsatisfactory results when it encounters visual content different from the past tasks (see Fig.~\ref{figure:shift_plot}). To tackle this, we propose to employ Test-Time Adaptation (TTA), which enables plasticity without suffering from CF. This TTA phase consists of a slight feature adaptation on the test images in an unsupervised fashion. After evaluation on each test batch, our model is reset to the initial state $E_{\theta,\phi}^*$ to maintain generalizability, and we prevent CF~\citep{fleuret2021test, zhang2021memo}. An additional reason for adopting a TTA approach is that it enhances the robustness of our model against noisy and corrupted samples during inference~\citep{sun2020test, wang2021tent}.
Inspired by MEMO~\citep{zhang2021memo}, given a test sample $\bx_i$, an adapted PTM model (from Phase I), and a set of image transformations $\augset$, we pick $M$ random transformations $\{\tau_{1}, \ldots, \tau_{M}\}$ from $\augset$ and apply them to $\bx_i$ to produce a batch of augmented data $\{\tilde{\bx}_{i,1},\ldots,\tilde{\bx}_{i,M}\}$. The marginal output distribution for the augmented points is given by:
\begin{equation}
    \bar{p}_{\theta, \phi}(y\vert\bx_i)\approx\frac{1}{M}\sum_{m=1}^Mp_{\theta, \phi}(y\vert\tilde{\bx}_{i,m}).
    \label{eq:marg}
\end{equation}
We propose to adapt the model by minimizing the entropy of its marginal output distribution over augmentations~\eqref{eq:marg}:
\begin{equation}
    \mathcal{L}_{\text{TTA}}(\theta, \phi;\bx_i)=-\sum_{y\in\outspace}\bar{p}_{\theta, \phi}(y\vert\bx_i)\log\bar{p}_{\theta, \phi}(y\vert\bx_i)\,.
    \label{eq:loss}
\end{equation}
Minimizing \eqref{eq:loss} encourages both high confidence and invariance to input augmentations.
\vspace{-3mm}
\begin{figure}[h]
    \centering
    \begin{minipage}{0.48\textwidth}
        Specifically, the entropy term $\bar{p}_{\theta, \phi}(\cdot\vert\bx_i)$ reaches its minimum value when the model delivers consistent and confident predictions, regardless of the particular data augmentation applied. The motivation for optimizing the model to yield more confident predictions is grounded in the belief that the decision boundaries distinguishing classes are situated in areas of the data space with lower density~\citep{grandvalet2004semi, zhang2021memo}.
        As $\bar{p}_{\theta, \phi}(y\vert\x_i)$ is differentiable with respect to $\params$ based on the loss defined in \eqref{eq:loss}, we can employ gradient-based optimization directly to adapt $\params$. Test-time adaptation requires careful selection of parameters~\citep{niu2022efficient}, our adaptation procedure focuses solely on adapting the \emph{layer-norm} parameters~\citep{ba2016layer}, unlike MEMO~\citep{zhang2021memo}, which finetunes the entire model. We found that adapting all the parameters hurts the performance (see Sec.~\ref{ablation_studies}). Only one gradient step per test point is empirically determined to be sufficient for achieving improved performance, with less inference time overhead.
    \end{minipage}%
    \hfill
    \begin{minipage}{0.48\textwidth}
        \centering
        \vspace{-5mm}
        \begin{algorithm}[H]
            \caption{\mame .}
            \label{alg:overall}
            \begin{algorithmic}[1]
                \REQUIRE{Sequence of $T$ tasks, $\mD\!={\mD^1,...,\mD^T}$, and pre-trained model $f$}
                \STATE \textbf{Phase I}: \emph{Adapting PTMs to New Tasks}
                \STATE \hskip1.0em Adapt model $f$ using Adapters on $\mD^1$.
                \STATE \textbf{Phase II:} \emph{Test-Time Model Refinement}
                \STATE{\textbf{Initiate} student/teacher models.}
                \FOR{a batch $\mX$ in $\mD_{test}^{t}$}
                    \FOR{iteration $n=1 \cdots N$}
                        \STATE Sample transformations from $\mathcal{T}$ and apply to $\mX$.
                        \STATE Calculate predictions via~\eqref{eq:marg}.
                        \STATE Adapt parameters via~\eqref{eq:total_loss}.
                    \ENDFOR
                    \STATE{\textbf{Update} the teacher model via \eqref{eq:ema_update}}.
                    \STATE{\textbf{Reset} the student model.}
                \ENDFOR
                \RETURN predictions for all $\bx_i \in \mD_{test}^{t}$. %$\{\hat{y}\}_{j=1}^M$ 
            \end{algorithmic}
        \end{algorithm}
    \end{minipage}
\end{figure}

\vspace{-5mm}
While TTA is a simple and effective solution for CIL, it does not fully conform to the tenets of continual learning as TTA involves model reset after each batch update \citep{zhang2021memo}, erasing all past updates. Moreover, although TTA can also be realized \textit{without} resetting its weights, such an approach will be prone to CF and reduced PTM generalization. To fully capitalize the past updates and prevent overfitting on the current batch (see Sec.~\ref{ablation_studies}), a regularization strategy is proposed, which is discussed next.

\noindent\textbf{Teacher-student distillation.} We propose to distill the past (and potentially useful) knowledge while performing TTA on a current test batch through a teacher-student distillation strategy.
Formally, the teacher model $E_{\theta_{T},\phi_{T}}$ is initially a copy of the student model before adaptation begins. Throughout the TTA phase, the student model is updated with each test batch as explained above. After each batch, the teacher model’s predictions are used to provide soft targets for the student model. These predictions are compared to the student model's outputs using Kullback-Leibler (KL) divergence~\citep{kullback1951information}. For simplification, we omit the parameters of Adapters $\phi$ from the equations, since they are frozen during Phase II. Thus, the KL-divergence loss is defined as follows:
\begin{equation}
\begin{aligned}
    \mathcal{L}_{KL}(\theta_T, \theta_S) &= \sum_{y \in \outspace} p_{\theta_S}(y \mid \tilde{\bx}_i) \log \frac{p_{\theta_S}(y \mid \tilde{\bx}_i)}{p_{\theta_T}(y \mid \tilde{\bx}_i)},
    \label{eq:kl_loss}
\end{aligned}
\end{equation}
where $p_{\theta_T}(y \mid \tilde{\bx}_i)$ and $p_{\theta_S}(y \mid \tilde{\bx}_i)$ represent the teacher and student model’s probability distributions over the class labels $y$, respectively, for the augmented input $\tilde{\bx}_i$. 

Overall, this teacher-student distillation approach helps in transferring the improved feature representation and decision boundaries learned by the teacher model to the student model, thereby benefiting from the TTA phase and enhancing the robustness of the final model. The overall loss during Phase II is given by:
\begin{equation}
    \mathcal{L}_{\text{total}}(\theta_S, \theta_T; \bx_i) = \mathcal{L}_{\text{TTA}}(\theta_S; \bx_i) + \lambda \mathcal{L}_{KL}(\theta_T, \theta_S),
    \label{eq:total_loss}
\end{equation}
where $\lambda$ is a hyperparameter that controls the weight of the KL-divergence loss relative to the TTA loss.

To ensure that the teacher model effectively incorporates the student’s learned information, we update the teacher using an Exponential Moving Average (EMA) of the student’s parameters after each test batch. EMA helps in transferring the learned representations, which would otherwise be lost in TTA. The EMA update is given by: %$\theta_T\!=\!\alpha\theta_T+(1\!-\!\alpha)\theta_S$
\begin{equation}
\theta_T = \alpha \theta_T + (1 - \alpha) \theta_S,
    \label{eq:ema_update}
\end{equation}
where $\theta_T$ and $\theta_S$ denote the teacher and student parameters, respectively, and $\alpha$ is the smoothing factor (with $0\!<\!\alpha\!<\!1$). Afterward, the student model is reset after each batch to incorporate effective adaptation on the test samples and avoid degradation of performance (see Sec.~\ref{ablation_studies}).
\section{Experiments}
\definecolor{lightgreen}{RGB}{200,255,200}
\begin{table*}[t]
    \centering
    \caption{\small Average (\(\bar{\mathcal{A}}\)) and last performance (\(\mathcal{A}_T\)) comparison on six datasets with {\bf ViT-B/16-IN21K} backbone.  ``IN-R/A'' stands for ``ImageNet-R/A'', and ``OmniBench'' stands for ``OmniBenchmark''. LwF, L2P, DualPrompt, and Coda-Prompt results are from~\citep{AdamAdapter}. The best performance is in bold, and the second best is underlined.}
    \vspace{-2mm}
    \resizebox{\textwidth}{!}{%
    \begin{tabular}{@{}c|l|cccccccccccc|c}
        \toprule
        \multirow{11}{*} & \multicolumn{1}{c}{\multirow{2}{*}{Method}} &
        \multicolumn{2}{c}{CIFAR100-Inc5} &
        \multicolumn{2}{c}{CUB-Inc10} &
        \multicolumn{2}{c}{IN-R-Inc5} &
        \multicolumn{2}{c}{IN-A-Inc10} &
        \multicolumn{2}{c}{OmniBench-Inc30} &
        \multicolumn{2}{c}{VTAB-Inc10} &
        \multicolumn{1}{c}{Average} \\
        & & {$\bar{\mathcal{A}}$} & ${\mathcal{A}_T}$ &
        {$\bar{\mathcal{A}}$} & ${\mathcal{A}_T}$ &
        {$\bar{\mathcal{A}}$} & ${\mathcal{A}_T}$ &
        {$\bar{\mathcal{A}}$} & ${\mathcal{A}_T}$ &
        {$\bar{\mathcal{A}}$} & ${\mathcal{A}_T}$ &
        {$\bar{\mathcal{A}}$} & ${\mathcal{A}_T}$ &
        {$\bar{\mathcal{A}}$} \\
        \midrule
        & NCM & 86.27 & 81.27 & 90.79 & 86.77 & 61.63 & 54.33 & 58.66 & 48.52 & 80.63 & 73.33 & 85.99 & 84.44 & 77.33 \\
        \midrule
        \multirow{6}{*}{\rotatebox{90}{\parbox{2cm}{Sequential \\ Updating}}} 
        & Finetune & 38.90 & 20.17 & 26.08 & 13.96 & 21.61 & 10.79 & 21.60 & 10.96 & 23.61 & 10.57 & 34.95 & 21.25 & 27.79 \\
        & Finetune Adapter ~\citep{chenadaptformer} & 60.51 & 49.32 & 66.84 & 52.99 & 47.59 & 40.28 & 43.05 & 37.66 & 62.32 & 50.53 & 48.91 & 45.12 & 54.87 \\
        & LwF ~\citep{li2017learning} & 46.29 & 41.07 & 48.97 & 32.03 & 39.93 & 26.47 & 35.39 & 23.83 & 47.14 & 33.95 & 40.48 & 27.54 & 43.03 \\
        & L2P~\citep{wang2022learning} & 85.94 & 79.93 & 67.05 & 56.25 & 66.53 & 59.22 & 47.16 & 38.48 & 73.36 & 64.69 & 77.11 & 77.10 & 69.53 \\
        & DualPrompt~\citep{wang2022dualprompt} & 87.87 & 81.15 & 77.47 & 66.54 & 63.31 & 55.22 & 52.56 & 42.68 & 73.92 & 65.52 & 83.36 & 81.23 & 73.08 \\

        & Coda-Prompt~\citep{CODAPrompt} & 89.11 & 81.96 & 84.00 & 73.37 & 64.42 & 55.08 & 53.54 & 42.73 & 77.03 & 68.09 & 83.90 & 83.02 & 75.33 \\
        
        & SLCA~\citep{SLCA} & \textbf{94.54} & \textbf{91.19} & 84.70 & 81.76 & 74.10 & 69.73 & 61.98 & 56.48 & 80.40 & 72.05 & 88.92 & 82.65 & 80.77 \\
        & EASE~\citep{zhou2024expandable} & 91.51 & 85.80 & \underline{91.01} & \underline{86.81} & \textbf{78.01} & \textbf{70.58} & \underline{65.80} & \underline{56.87} & \underline{81.03} & \underline{74.85} & \textbf{92.61} & \textbf{90.55} & \underline{83.32} \\
        \midrule
        \multirow{5}{*}{\rotatebox{90}{\parbox{2cm}{First Session \\ Adaptation}}} & FSA-FiLM~\citep{panos2023first} & 90.13 & 86.23 & 91.23 & 86.12 & 73.65 & 65.13 & 60.11 & 49.53 & 80.64 & 74.09 & 87.47 & 82.12 & 80.53 \\
        %& \adam$\dagger$~\citep{AdamAdapter} & 90.65 & 85.15 & 92.21 & 86.73 & 72.35 & 64.33 & 60.53 & 49.57 & 80.75 & 74.37 & 85.95 & 84.35 & 80.41 \\
        & \adam~\citep{AdamAdapter} & 90.87 & 85.15 & 91.06 & 86.73 & 72.35 & 64.33 & 60.21 & 49.57 & 79.70 & 55.24 & 85.95 & 84.35 & 80.02 \\
        & RanPAC~\citep{ranpac} & \underline{93.14} & \underline{89.05} & 91.56 & 85.91 & 77.60 & 68.03 & 61.84 & 46.02 & 80.27 & 74.46 & \underline{91.75} & \underline{90.32} & 82.69  \\
        & FeCAM~\citep{fecam} & 88.97 &  \imad{85.34} &  \imad{\textbf{92.16}} &  \imad{88.59} &  \imad{70.13} &  \imad{65.17} &  \imad{56.78} &  \imad{47.27} & \imad{\textbf{83.34}} & \imad{73.03} &  \imad{\textbf{94.14}} & \imad{89.12} & \imad{80.92} \\
        
        \cmidrule{2-15}
        & \cellcolor{lightgreen}\method (w/o Phase I) & \cellcolor{lightgreen}85.71& \cellcolor{lightgreen}82.54 &\cellcolor{lightgreen}89.55& \cellcolor{lightgreen}85.75 &\cellcolor{lightgreen}61.96 & \cellcolor{lightgreen}55.13 & \cellcolor{lightgreen}60.46 & \cellcolor{lightgreen}50.43 & \cellcolor{lightgreen}81.05& \cellcolor{lightgreen}73.84 & \cellcolor{lightgreen}89.93 & \cellcolor{lightgreen}85.17 & \cellcolor{lightgreen}78.11 \\
        
        & \cellcolor{lightgreen}\name & \cellcolor{lightgreen}92.34 & \cellcolor{lightgreen}88.58 & \cellcolor{lightgreen}\textbf{91.24}& \cellcolor{lightgreen}\textbf{86.95} & \cellcolor{lightgreen}\underline{76.88} & \cellcolor{lightgreen}\underline{69.93} & \cellcolor{lightgreen}\textbf{67.04} & \cellcolor{lightgreen}\textbf{57.41} & \cellcolor{lightgreen}\textbf{81.13} & \cellcolor{lightgreen}\textbf{76.39} & \cellcolor{lightgreen} 91.57 & \cellcolor{lightgreen}86.38 & \cellcolor{lightgreen}\textbf{83.36} \\
        \bottomrule
    \end{tabular}
    }
    \vspace{-3mm}
    \label{tab:benchmark}
\end{table*}

\begin{table*}[t]
    \centering
    \caption{\small Average performance (\(\bar{\mathcal{A}}\)) comparison with {\bf ViT-B/16-IN21K} as the backbone on three corrupted benchmarks CIFAR100-C, VTAB-C and OmniBench-C with Level-5 noise corruptions~\citep{hendrycks2019benchmarking}. The best performance is in bold, and the second best is underlined.}
    \vspace{-2mm}
   \label{benchmarkcorrupted}
		\resizebox{0.95\textwidth}{!}{%
		\begin{tabular}{l|cccc|cccc|cccc}
  
			\toprule
			\multicolumn{1}{l}{\multirow{2}{*}{Method}} & 
			\multicolumn{4}{c}{CIFAR100-C} & 
                \multicolumn{4}{c}{VTAB-C} & 
			\multicolumn{4}{c}{OmniBench-C} \\
                & Gauss. & Shot & Impul. & Avg.
                & Gauss. & Shot & Impul. & Avg.
                & Gauss. & Shot & Impul. & Avg. \\
		\midrule
        NCM	& 35.52&	39.05&	46.89& 40.48	& 69.64 &	72.31&	58.18 &  66.71 & 77.44&	78.43&	75.07 & 76.98\\
	\adam~\citep{AdamAdapter}   
        & 37.52&	48.73&	65.03& \underline{50.43} & 71.10&	74.11&	58.98 & \underline{68.06} & 79.45&79.57&	77.76 & \underline{78.93} \\
        \midrule
        \rowcolor{lightgreen}
	\name & 48.87&	53.38&	68.61&	\textbf{56.95} & 71.58	& 74.23 & 59.44 &  \textbf{68.41} & 79.70 & 79.81 & 78.41 & \textbf{79.31}\\
			\bottomrule
		\end{tabular}%
	}
   \vspace{-3mm}
\end{table*}

\subsection{Experimental setup}
\noindent {\bf Datasets and settings.} We validate \method on seven CIL benchmarks. In detail, following the work in~\citep{AdamAdapter} we experiment with CIFAR100~\citep{krizhevsky2009learning}, CUB200~\citep{WahCUB2002011}, ImageNet-R~\citep{hendrycks2021many}, ImageNet-A~\citep{hendrycks2021natural}, OmniBench~\citep{zhang2022benchmarking}, 5-Datasets~\citep{wang2022learning} and VTAB~\citep{AdamAdapter}. All these benchmarks offer diverse levels of difficulty in terms of high task heterogeneity (\eg 5-Datasets, VTAB) or large domain gap to ImageNet (\eg ImageNet-A). For the experiments on robustness, we pick three benchmarks: CIFAR100-C~\citep{hendrycks2019benchmarking}, OmniBench-C, and VTAB-C, where ``C'' stands for corrupted counterparts of the original benchmarks. In this work, we introduce the VTAB-C and OmniBench-C by adding the same corruptions as in CIFAR100-C to the original testing sets of VTAB and OmniBench benchmarks, respectively.

We follow the incremental settings established in~\citep{AdamAdapter}, \ie by adopting an increment per task of three types: 5 classes (Inc5) for CIFAR100 and ImageNet-R; 10 classes (Inc10) for CUB200, ImageNet-A, VTAB, and 5-Datasets; and 30 classes (Inc30) for OmniBench. The same increments and task orders are followed in the robustness experiments. Finally, we evaluate all the methods in \textit{task-agnostic} manner, \ie, without having access to the task-id at inference. More details are available in the Supp. Mat.

\noindent\textbf{Implementation details.} Following \adam~\citep{AdamAdapter}, we have used ViT-B/16~\citep{dosovitskiy2021image} initially pre-trained on ImageNet-21K and further fine-tuned on ImageNet-1K, as a backbone. For Phase I, we trained the adapters (one adapter per ViT block) using Stochastic Gradient Descent (SGD) with momentum for 20 epochs, with an initial learning rate of 0.01 that undergoes cosine annealing. In Phase II we have used a batch size of 16 and augmented each test sample 8 times using random image transformations~\citep{zhang2021memo}. As mentioned before, for one mini-batch, we do one gradient step in Phase II to obtain the adapted model. The learning rate is set to 0.01, no weight decay is employed, $\lambda\!=\!0.1$, and $\alpha\!=0.99$. We use PILOT~\citep{sun2023pilot} framework for our experiments. 

%\vspace{2mm}
\noindent {\bf Baselines and competitors.} We have compared \method with state-of-the-art PTM-based CIL methods: LwF~\citep{kirkpatrick2017overcoming}, L2P~\citep{wang2022learning}, DualPrompt~\citep{wang2022dualprompt}, \adam\citep{AdamAdapter}, FSA-FiLM~\citep{panos2023first}, EASE~\citep{zhou2024expandable}, \imad{FeCAM~\citep{fecam}} and SLCA~\citep{SLCA}. All methods are initialized with the same PTM (ViT-B-16~\citep{dosovitskiy2021image}) pre-trained on Imagenet-21K. Additionally, we have compared with two more baselines: Finetune and Finetune Adapter~\citep{chenadaptformer}.

\noindent \textbf{Evaluation Metrics.} In alignment with ~\citep{AdamAdapter, zhou2024expandable}, %we denote the Top-1 accuracy achieved at the end of the \(t\)-th task as \(\mathcal{A}_t\).
we employ two key performance indicators: (i) \(\mathcal{A}_T\), representing the Top-1 accuracy after learning the final task, and (ii) \(\bar{\mathcal{A}}\), calculated as the average Top-1 accuracy while learning progresses in incremental steps. \imad{Additionally, we report the standard deviation for both metrics computed over five independent trials (seeds).}
\vspace{-5mm}
\subsection{Main results}
\label{sec:mainresults}
\vspace{-3mm}
\noindent\textbf{Results on standard CIL benchmarks. }We report in Tab.~\ref{tab:benchmark} the empirical evaluation on the standard CIL benchmarks. The results show that \method achieves the best performance, with an average accuracy \(\bar{\mathcal{A}}\) of 83.36\%, the highest among all methods. Notably, \method consistently outperforms \adam~\citep{AdamAdapter} and RanPAC~\citep{ranpac} on most benchmarks, including challenging datasets like ImageNet-A and OmniBench. While SLCA and EASE slightly outperform \method on specific datasets like CIFAR100 and VTAB, their competitive results stem from their ability to train significantly more parameters—SLCA fine-tunes all model parameters, and EASE introduces a new adapter for each task. This increased plasticity comes at the cost of substantial computational overhead, making these methods less efficient and scalable compared to \method. Compared to \adam, \method shows a substantial improvement, with a gap of +3.34\% in average performance (83.36\% vs 80.02\%). If adapters are fine-tuned on each task, as seen in the Finetune Adapter baseline, the final performance drops drastically to 54.87\%, highlighting the severe downsides of task-specific fine-tuning. \method w/o Phase I achieves competitive results with an average accuracy of 78.11\%, which is still a notable improvement over the baseline NCM (+0.78\%). %Particularly, on VTAB, \method (w/o Phase I) outperforms NCM by +3.94\%, underscoring the limitations of frozen PTMs for datasets with different distributions from pre-training data.

\noindent \textbf{Results on CIL benchmarks with corruptions.} In Tab.~\ref{benchmarkcorrupted} we report the empirical evaluation on CIL benchmarks with different kinds of noise corruptions (of level 5~\citep{hendrycks2019benchmarking}), where we compare our proposed \method with the competitors. In \adam and our \method, \textit{we exclude training on the clean samples from every task, except the first task}. The NCM exhibits the worst performance as it solely relies on the representation learned from the pre-training dataset. \adam, which trains the network on the training data from the first task, shows an improvement in performance due to better adaptivity to the NCM. \method achieves the best performance as it further allows the model to adapt towards every subsequent task. In particular, the gain in performance is significant for CIFAR100-C, where our proposed \method outperforms \adam by +5.33\%.\\
\indent In summary, \method satisfies three desiderata: (i) improved adaptivity to new tasks; (ii) less computational cost by avoiding sequential training on each task; and (iii) robustness to corruptions by refining its predictions on the corrupted test instances during test-time. We report the results on other types of corruption in the Supp. Mat.
\vspace{-2mm}
\subsection{Ablation study}
\label{ablation_studies}
\begin{figure*}[t]
    \centering
    \vspace{-5mm}
    \begin{minipage}{0.48\textwidth}
        \centering
        \includegraphics[width=\linewidth]{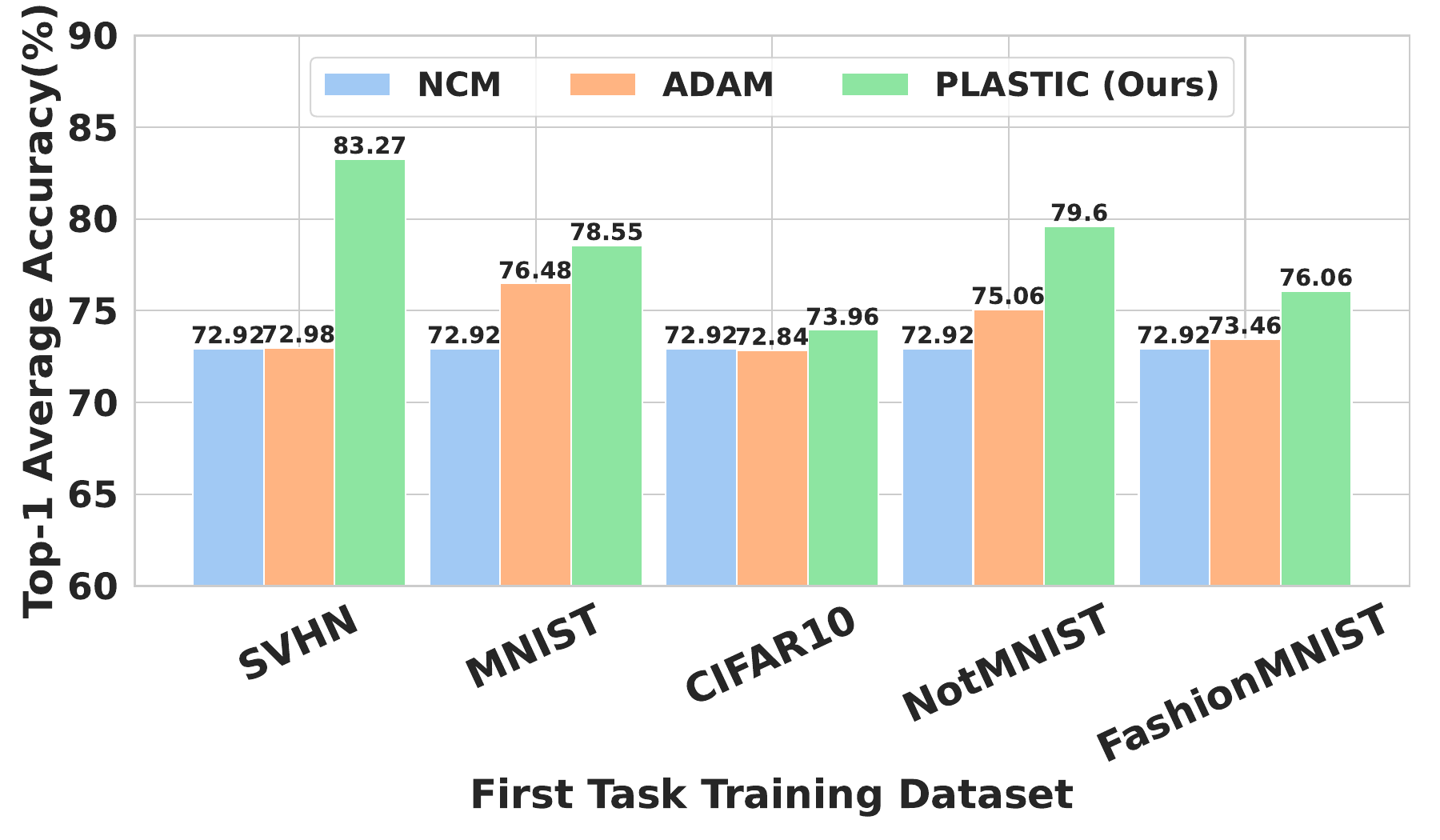}
        \vspace{-6mm}
        \caption{Impact of initial task on model performance. X-axis: first training task. Y-axis: average performance across datasets for NCM, \adam, and \method. Performance varies with task order, highlighting the importance of the initial task.}
        \label{figure:shift_plot}
    \end{minipage}%
    \hfill
    \begin{minipage}{0.48\textwidth}
    \vspace{-14mm}
        \centering
        \includegraphics[width=\linewidth]{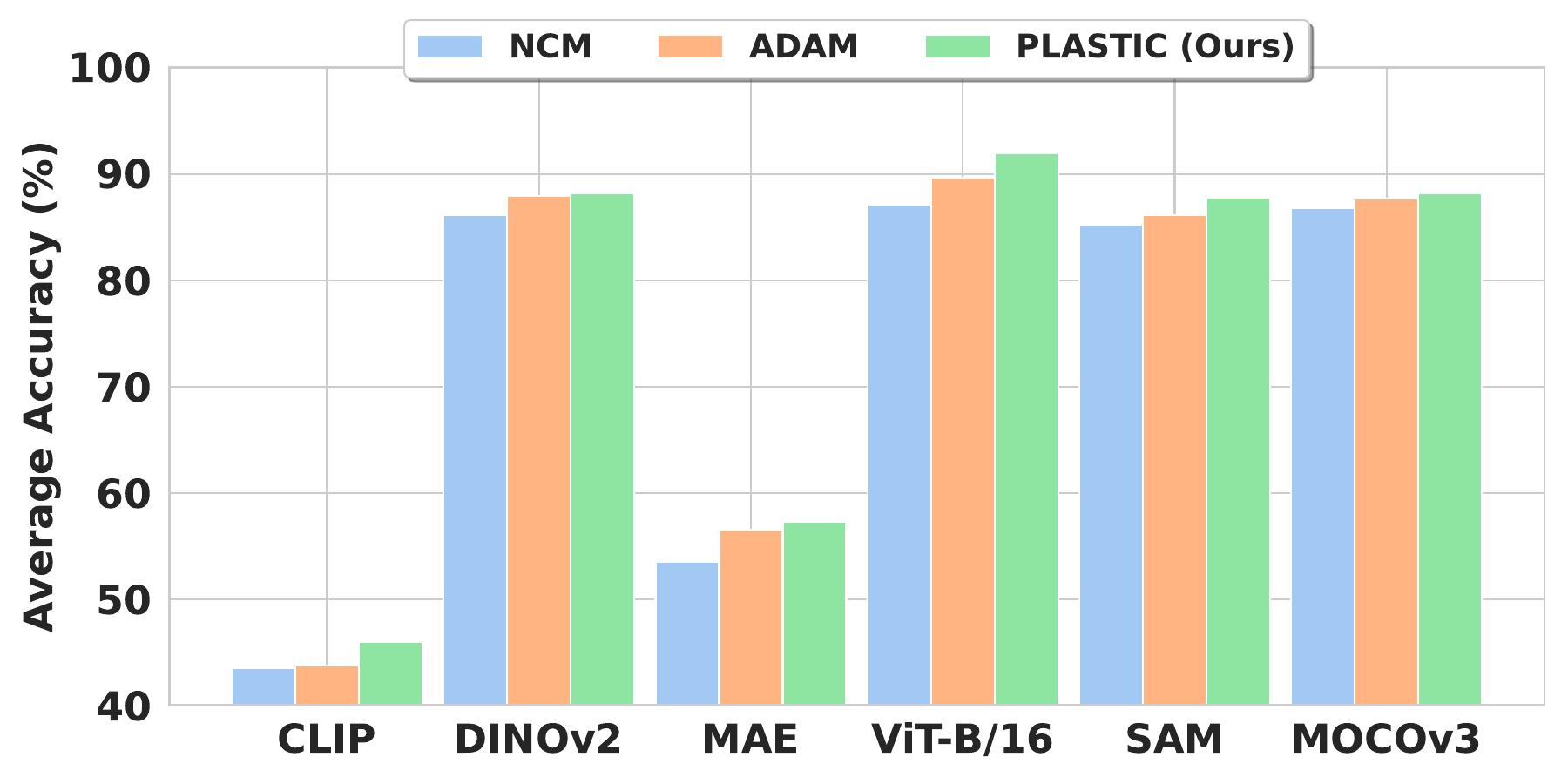}
        \caption{Evaluation with different PTMs on VTAB. \mame{} consistently improves the performance of ADAM and NCM.}
        \label{figure:ptms_ablation}
    \end{minipage}
\end{figure*}

\vspace{-2mm}
\noindent \circled{1} \textbf{Effect of task ordering.} In Fig.~\ref{figure:shift_plot} we study the influence of task order, especially in the case of heterogeneous tasks. We choose to perform experiments on the 5-Datasets benchmark~\citep{wang2022dualprompt} as it contains five very heterogeneous datasets: SVHN, MNIST, CIFAR10, NotMNIST, and FashionMNIST.
Here we mainly compare \method with \adam as both undergo training on the first task's data. From Fig.~\ref{figure:shift_plot}, we notice that a different task order yields different performance for \method and \adam, indicating that the first task plays an important role in determining the final performance. However, in all the task orderings, \method outperforms \adam by big margins, with the highest margin (+10.29\%) obtained in the tasks of order SVHN $\rightarrow$ MNIST $\rightarrow$ CIFAR10 $\rightarrow$ NotMNIST $\rightarrow$ FashionMNIST. This highlights that the data from the second task onwards, despite \textit{unlabelled}, offer a learning signal that is efficiently exploited by our \method during Phase II.
Differently, as \adam does not offer any plasticity after the first task, it fails to improve the adaptivity on future tasks.

We can also observe that when the first task is CIFAR10, a dataset very different from the other tasks, \method still yields a performance gain over both \adam and NCM. These results are promising since they indicate that \method is a relevant solution even when the domain shifts in future tasks are potentially large~\citep{prabhu2023computationally}.

\vspace{-5mm}
\begin{table}[H]
    \centering
    \renewcommand{\arraystretch}{1.2} % Improve row spacing
    \begin{minipage}{0.48\textwidth} % Text on the right
        \noindent \circled{2} \textbf{Effect of the adaptation parameters. }\label{ablation_params}
        The experiments presented in Tab.~\ref{tab:ablation_compo} focus on evaluating different parameters to adapt, and the effect of model resetting during \emph{Test-Time Model Refinement phase}. We compare: (i) \textit{no TTA} -- we do not apply test-time adaptation (equivalent to ADAM~\citep{AdamAdapter}); (ii) \textit{All} -- we adapt all the parameters of the model; (iii) \textit{Adapter} -- we adapt only the parameters of the adapter; and (iv) \textit{Norm} -- we adapt only the Layer Norm.
    \end{minipage}
    \hfill
    \begin{minipage}{0.48\textwidth} % Table on the left
        \centering
        \caption{Performance of \method with various optimized parameters during Phase II. M: Trainable parameters in millions ($\downarrow$).}
        \vspace{-2mm}
        \label{tab:ablation_compo}
        \resizebox{\textwidth}{!}{
        \begin{tabular}{cc|c|cc}
            \toprule
            & & \# Par. (M) $\downarrow$ & VTAB ($\bar{\mathcal{A}}$) $\uparrow$ & ImageNet-R ($\bar{\mathcal{A}}$) $\uparrow$ \\
            \midrule
            & No TTA & 0 & $85.68 \pm 2.44$ & $72.35 \pm 1.11$ \\
            \midrule
            \multirow{3}{*}{\rotatebox{90}{TTA}}
            & All & 87 & $43.26 \pm 1.15$ & $26.02 \pm 0.39$ \\
            & Adapter & 1.2 & $89.89 \pm 0.64$ & $75.28 \pm 0.97$ \\
            & Norm & \textbf{0.04} & \textbf{91.57} $\pm$ 0.64 & \textbf{76.88} $\pm$ 0.42 \\
            \bottomrule
        \end{tabular}
        }
    \end{minipage}%
    \vspace{-5mm}
\end{table}
\vspace{-2mm}
Adapting all the parameters results in 87M trainable parameters, yielding lower VTAB and Imagenet-R accuracies of 43.26\% and 26.02\% respectively, likely due to the inherent complexity of adapting with few samples, and to one optimization iteration. The adapter-only update reduces trainable parameters to 1.2M, improving VTAB and Imagenet-R accuracies to 89.89\% and 75.28\%. The most efficient performance is achieved by updating only 0.04M Layer Norm parameters, aligning with Niu~\etal's~\citep{niu2023towards} findings on norm-layer adaptations enhancing stability in TTA. Further, our approach achieves superior efficiency by requiring significantly fewer trainable parameters while maintaining high performance. We provide a computational complexity analysis and a comparison of trainable parameters against other baselines in the supplementary material, highlighting this advantage.

\begin{table}[H]
    \centering
    \renewcommand{\arraystretch}{1.2} % Improve row spacing
    \begin{minipage}{0.48\textwidth} % Text on the right
    \noindent \circled{3} \textbf{Effect of Student-Teacher distillation.} Tab.~\ref{tab:ablation_no_tta} \imad{presents a study analyzing the contribution of two core components in the PLASTIC: the \textit{student reset} and the \textit{KL divergence loss} (\eqref{eq:kl_loss}) used for distillation between student and teacher. Without student resetting, performance drops significantly—down to 43.31\% on VTAB and 41.77\% on ImageNet-A—due to cumulative overfitting and knowledge degradation~\citep{zhao2023pitfalls}.}
    \end{minipage}
    \hfill
    \vspace{-2mm}
    \begin{minipage}{0.48\textwidth}
        \centering
        \caption{Impact of distillation framework components.}
        \vspace{-3mm}
        \label{tab:ablation_no_tta}
        \resizebox{\textwidth}{!}{
        \begin{tabular}{cc|cc}
            \toprule
            Reset & $\mathcal{L}_{KL}$ & VTAB ($\bar{\mathcal{A}}$) $\uparrow$ & ImageNet-A ($\bar{\mathcal{A}}$) $\uparrow$ \\
            \midrule
             & & $43.31 \pm 7.34$ & $41.77 \pm 9.29$ \\
             & \checkmark & $59.15 \pm 3.85$ & $43.12 \pm 4.08$ \\
            \checkmark & \checkmark & \textbf{91.57} $\pm$ 0.64 & \textbf{67.04} $\pm$ 0.53 \\
            \bottomrule
        \end{tabular}
        }
    \end{minipage}%
\end{table}
\vspace{-5mm}
\imad{Resetting the student to the adapted pre-trained model mitigates this drift by reinitializing adaptation from a model with strong generalization ability, thereby reducing variance and overfitting. However, resetting alone is not sufficient, as the student—despite being reinitialized—can still overfit to the current batch during adaptation. To mitigate this, we incorporate a KL divergence loss from the teacher to the student. This loss acts as a regularization signal, guiding the student to adapt while remaining close to the teacher’s stable representation, which encapsulates previously accumulated knowledge. In parallel, the teacher model is updated via an EMA of the student, allowing it to gradually integrate useful information across batches without being overly influenced by any single adaptation step. Together, the KL loss and reset mechanism ensure that adaptation is stable and knowledge accumulation.} 
\vspace{-3mm}
\begin{table}[H]
    \centering
    \renewcommand{\arraystretch}{1.2} % Improve row spacing
    \begin{minipage}{0.48\textwidth} % Adjust width for text
        \noindent \circled{4} \textbf{Effect of TTA. }Our analysis proposed in Tab.~\ref{tab:ablation_tta_methods} focuses on the effectiveness of different TTA methods—specifically TENT~\citep{wang2021tent}, MEMO~\citep{zhang2021memo}, and SAR~\citep{niu2023towards}— in \method while constraining fine-tuning to layer-norm~\citep{ba2016layer}.
    \end{minipage}%
    \hfill % Ensures spacing between the text and the table
    \begin{minipage}{0.48\textwidth}
        \centering
        \caption{Evaluation of TTA methods in \method.}% Average performance ($\bar{\mathcal{A}}$) is reported for VTAB and ImageNet-A (higher is better $\uparrow$).
        \vspace{-3mm}
        \label{tab:ablation_tta_methods}
        \resizebox{\textwidth}{!}{
        \begin{tabular}{c|cc}
            \toprule
            Method & VTAB ($\bar{\mathcal{A}}$) $\uparrow$ & ImageNet-A ($\bar{\mathcal{A}}$) $\uparrow$ \\
            \midrule
            TENT~\citep{wang2021tent} & $81.21 \pm 0.87$ & $61.40 \pm 0.70$ \\
            SAR~\citep{niu2023towards} & $83.79 \pm 1.15$ & $55.69 \pm 1.92$ \\
            MEMO~\citep{zhang2021memo} & \textbf{91.57} $\pm$ 0.64 & \textbf{67.04} $\pm$ 0.53 \\
            \bottomrule
        \end{tabular}
        }
    \end{minipage}%
\end{table}
\vspace{-7mm}
The evaluation of TTA methods in \method reveals distinct performance trends: MEMO consistently outperforms TENT and SAR, achieving the highest VTAB (91.28\%) and ImageNet-A (67.04\%) accuracies with minimal variance. SAR surpasses TENT on VTAB but exhibits higher variability. These results suggest that TENT and SAR's effectiveness depends on dataset characteristics and task requirements~\citep{zhao2023pitfalls}.
\vspace{-3mm}
\begin{table}[H]
    \centering
    \renewcommand{\arraystretch}{1.2} % Improve row spacing
    \begin{minipage}{0.55\textwidth} % Adjust width for text
        \noindent \circled{5} \textbf{Effect of the test samples order. }We assess \method's sensitivity to test sample order, a critical factor in real-world applications. Tab.~\ref{tab:tta_order} reports results from five random seeds on VTAB, ImageNet-R, and ImageNet-A. \method demonstrates strong robustness, maintaining consistently high performance with low variance across datasets. Notably, VTAB results show exceptional stability ($90.60 \pm 1.87$ in $\bar{\mathcal{A}}$, $86.02 \pm 0.31$ in ${\mathcal{A}_T}$). 
    \end{minipage}%
    \hfill % Ensures spacing between the text and the table
    \begin{minipage}{0.42\textwidth} % Adjust width for table
        \centering
        \caption{Robustness of \method to test sample order across datasets.}%Results with low standard deviation ($\pm$) demonstrate consistent performance regardless of test sample sequence.
        \label{tab:tta_order}
        \vspace{-3mm}
        \resizebox{\linewidth}{!}{
        \begin{tabular}{c|cc}
            \toprule
            \multirow{1}{*}{Method} &
            \multirow{1}{*}{$\bar{\mathcal{A}}$ $\uparrow$} &
            \multirow{1}{*}{${\mathcal{A}_T}$ $\uparrow$} \\
            \midrule
            VTAB & 90.60 $\pm$ 1.87 & 86.02 $\pm$ 0.31 \\
            ImageNet-A & 66.46 $\pm$ 1.37 & 55.72 $\pm$ 1.29 \\
            ImageNet-R & 76.62 $\pm$ 0.59 & 70.66 $\pm$ 0.13 \\
            \bottomrule
        \end{tabular}  
        }
    \end{minipage}
\end{table}
\vspace{-7mm}
The consistency across seeds indicates that \method effectively learns from the test distribution without overfitting to specific batch orders. We attribute this robustness to our adapted student-teacher distillation framework, which facilitates continuous learning while mitigating overfitting to temporary distribution shifts.

\noindent \circled{6} \textbf{Effect of the PTM backbone. }Here we evaluate, on VTAB, the influence of pre-trained models (PTMs). The analysis in in Fig.~\ref{figure:ptms_ablation} incorporates PTMs such as ViT-B/16-IN1K/21K, ViT-B/16-DINO~\citep{{caron2021emerging}}, ViT-B/16-SAM~\citep{{chen2022vision}}, ViT-B/16-MAE~\citep{{he2022masked}}, and ViT-B/16-CLIP~\citep{{radford2021learning}} (image encoder). ViTs trained with supervised loss demonstrate enhanced performance over unsupervised counterparts. CLIP's performance on the VTAB benchmark is the lowest, as indicated by Radford~\etal~\citep{radford2021learning}. CLIP features for VTAB dataset subsets show reduced discriminability, leading to lower performance compared to alternative backbones. Furthermore, the nature of the pre-training task contributes to MAE's modest results on the VTAB benchmark. In summary, \mame{} surpasses ADAM across all backbones, attributed to the improvement during TTA.
\begin{table}[H]
    \centering
    \renewcommand{\arraystretch}{1.2} % Improve row spacing
    \begin{minipage}{0.55\textwidth}
    \noindent \circled{7} \textbf{Effect of the hyperparameters. }We investigate the influence of iterative adjustments during the \textit{test-time model refinement} phase. The results are summarized in Tab.~\ref{tab_iter}. Increasing the number of test-time iterations resulted in a decrease in performance on both datasets, due to overfitting on low-entropy samples, thereby impairing the model's generalization~\citep{niu2022efficient,zhao2023pitfalls}. Overfitting to low-entropy samples leads to inaccuracies in classifying more complex or higher-entropy samples within the same batch, highlighting a trade-off 
    \end{minipage}%
    \hfill % Ensures spacing between the text and the table
    \begin{minipage}{0.44\textwidth}
        \centering
        \vspace{-4mm}
        \caption{\imad{Effect of adaptation iterations $N$.}}
        \label{tab_iter}
        \vspace{-3mm}
        \resizebox{\linewidth}{!}{
        \begin{tabular}{c|cc}
            \toprule
            Iterations & VTAB ($\bar{\mathcal{A}}$) $\uparrow$ & Imagenet-R ($\bar{\mathcal{A}}$)$\uparrow$ \\
            \midrule
            $N=1$ & \textbf{91.28} $\pm$ 0.64 & \textbf{76.88} $\pm$ 1.02 \\
            $N=2$ & \underline{87.34} $\pm$ 3.05 & \underline{74.77} $\pm$ 1.22 \\ 
            $N=3$ & 84.07 $\pm$ 3.53 & 70.14 $\pm$ 3.35 \\
            $N=4$ & 80.78 $\pm$ 5.45 & 69.13 $\pm$ 2.08 \\
            \bottomrule
        \end{tabular}
        }
    \end{minipage}
\end{table}
\vspace{-6mm}
between adaptability and overfitting. We also investigated the influence of batch size, number of adaptation iterations, and augmentations per iteration on model performance (see Supp. Mat.).
\subsection{Discussion}
\vspace{-1mm}
\noindent \circled{1} \textbf{Computational complexity.} The main limitations of \method are its computational complexity and its high inference time. Phase II requires extra computation before each prediction, but this increased computation cost is a trade-off for improved plasticity and robustness. Practically, \mame{} can still perform inference at 8 frames per second when using a single TTA iteration (see Supp. Mat.). Furthermore, this increase in inference time is balanced by \mame{}'s computational efficiency during training, \ie, by updating only the Adapter parameters in the first task.

\noindent \circled{2} \textbf{Practical benefits.} The additional test-time overhead of \mame{} is also balanced by the greater computational efficiency in the training phase, necessitating parameter updates solely on the first task. This contrasts with conventional CIL strategies, which require fine-tuning of PTMs across all tasks~\citep{SLCA}. Therefore, \mame{} leads to a reduction in communication costs for the CIL model to multiple clients in a distributed inference system~\citep{9286189,9155237}.
\section{Conclusions and Future Work}
We introduced \method, a novel approach to CIL that integrates the stability of First Session Adaptation with the plasticity of Test-Time Adaptation. \method effectively adapts to both homogeneous and heterogeneous task distributions while preserving the generalization capabilities of pre-trained models. Our method achieves state-of-the-art performance on standard CIL benchmarks, demonstrating enhanced plasticity and inherent robustness to data corruptions. Future research will focus on further improving \method's efficiency and scalability, particularly by leveraging advanced sampling strategies~\citep{niu2022efficient, wang2023feature} to improve inference time.

\imad{\textbf{Acknowledgements. }This paper has been supported by the French National Research Agency (ANR) in the framework of its JCJC, and was funded by the European Union's Horizon Europe research and innovation program under grant agreement No. 101120237 (ELIAS). Furthermore, this research was partially funded by Hi!PARIS Center on Data Analytics and Artificial Intelligence. This work was also supported in part by project SERICS (PE00000014) under the NRRP MUR program funded by the EU - NGEU. This work was granted access to the HPC resources of IDRIS under the allocation AD011013860 made by GENCI.}

\bibliography{collas2025_conference}
\bibliographystyle{collas2025_conference}

% \appendix
% \section{Appendix}
\clearpage
\appendix
\setcounter{page}{1}

\imad{In this supplementary material, we provide more details about the experimental results mentioned in the main paper, as well as additional empirical evaluations and discussions. The supplementary material is organized as follows: In Section~\ref{sec:proto}, we provide provide more details about NCM. In Section~\ref{sec:TaskOrder}, we evaluate the effect of task order. In Section~\ref{sec:corruptions}, provide additional evaluations on CIFAR100-C dataset. In Section~\ref{sec:ablations}, reports the full experimental results ablation studies in the main paper. In Section~\ref{sec:inference}, we analyze the inference cost of \mame{}. In Section~\ref{sec:adapter}, provides an ablation study about the Adapter size in \mame{}. In Section~\ref{sec:baselines}, we detail the baselines, and datasets used in our experiments.}

\section{NCM with PTMs} 
\label{sec:proto}
Inspired by \citep{rebuffi2017icarl, mensink2013distance}, we use a NCM approach to transfer PTM knowledge for incremental tasks. Let $E_{\theta}$ be the PTM model's feature extractor. We compute class prototypes $\boldsymbol{c}_k$ for $k\!=\!1,...,K$ by averaging the embeddings of each class in the current task $\mD^t$ as:
\begin{equation}
    \boldsymbol{c}_k=\frac{1}{N_k}\sum_{j=1}^{|\mathcal{{D}}^t|}\delta_{y_j,k} E_{\theta}(\bx_j),
\label{eq:prototype}
\end{equation}
where $\delta_{a,b}$ is the Kronecker delta. For a test instance $\bx_i$, the class probability $p_{\theta}(y_i=k|\boldsymbol{z}_i)$ is computed using the dot product between the feature representation $\boldsymbol{z}_i\!=\!E_{\theta}(\bx_i)$ and each prototype:
\begin{equation}
\label{eq:argmax}
    p_{\theta} \left(y_i=k|\boldsymbol{z}_i\right)
= \frac{\exp(\boldsymbol{z}_i \cdot \boldsymbol{c}_{k})}{\sum_{j}\exp(\boldsymbol{z}_i \cdot \boldsymbol{c}_{j})}.
\end{equation}

Our model $f$, combining $E_{\theta}$ and prototype classifiers, achieves competitive performance without fine-tuning (see Sec.~\ref{sec:mainresults}). However, performance drops when downstream datasets exhibit domain shifts, especially with significant concept drift~\citep{alfassy2022feta, hendrycks2021natural}.

\section{Computational Analysis}
\label{comput_appendix}
\noindent Table~\ref{tab:complexity_tab} presents a comparative analysis of Class-Incremental Learning (CIL) methods across average accuracy, space complexity, and time complexity. \method (Ours) achieves the highest accuracy ($83.36\%$), slightly outperforming EASE ($83.32\%$) and RanPAC ($82.69\%$).

Regarding space complexity, \method and RanPAC maintain constant scaling (\(n + p\)) for storage. These methods train only on the first task, requiring no additional parameters for subsequent tasks. In contrast, EASE, which trains an adapter for each task, scales linearly with the number of tasks (\(n+p \cdot T\)). This method necessitates concatenating adapters during inference to update prototypes, increasing computational overhead. SLCA updates all model parameters, resulting in space complexity of \(n\).

In terms of time complexity, \method and RanPAC achieve task-independent training and inference complexities of \(\mathcal{O}(1)\), making them computationally efficient. Sequentially updated methods like EASE, SLCA, and Finetune Adapter incur higher training time complexities (\(\mathcal{O}(T)\)) due to the need to process all tasks incrementally. EASE further incurs task-dependent inference complexity (\(\mathcal{O}(T)\)) due to the concatenation of task-specific adapters. The analysis provided here focuses on whether inference complexity scales with the number of tasks; task-specific analysis subject to TTA in our \method is detailed in Section \ref{sec:inference}.

\method and RanPAC are parameter-efficient, introducing only $1\!-\!3\%$ additional trainable parameters through adapters. In contrast, EASE requires $1\!-\!3\%$ parameters per task, leading to a total proportional to the number of tasks. SLCA updates $100\%$ of the model parameters, significantly increasing memory requirements.

\method's combination of task-independent time complexity, constant space complexity, and high accuracy establishes it as a computationally and memory-efficient solution for CIL.

\begin{table}[H]
    \centering
    \caption{Comparison of class-incremental learning (CIL) methods based on average accuracy, space complexity (storage), and time complexity (training and inference) as functions of the number of tasks $T$. $n$ and $p$ represent the number of model's and adapter's parameters, respectively. Space complexity (storage) indicates the network size growth with additional tasks. Time complexity measures the computational cost during training and inference. NCM does not require training, while EASE, SLCA, and Finetune Adapter train sequentially on all tasks. \method and RanPAC train only on the first task. \method achieves the highest average accuracy (83.36\%) with constant space complexity $n + p$ and task-independent time complexity $\mathcal{O}(1)$.}

    \resizebox{1.0\linewidth}{!}{
    \begin{tabular}{cc|c|c|ccccc}
        \toprule
         & \multirow{2}{*}{\textbf{Methods}} & \multirow{2}{*}{\textbf{Trainable Param.}} & \textbf{Space Complexity} & \multicolumn{2}{c}{\textbf{Time Complexity}} & \multirow{2}{*}{\textbf{Avg. Accuracy}} \\
         \cline{5-6}
         & & & Storage & Training & Inference & \\
         \midrule
         \multirow{1}{*}{} & \textbf{NCM} & 0 & $n$ & $-$ & $\mathcal{O}(1)$ & 77.33 \\
         \midrule
         \multirow{3}{*}{Sequential Updating} 
         & \textbf{Finetune Adapter} & $1\!-\!3\%$ & $n+p$ & $\mathcal{O}(T)$ & $\mathcal{O}(1)$ & 54.87 \\
         & \textbf{SLCA \citep{SLCA}} & $100\%$ & $n$ & $\mathcal{O}(T)$ & $\mathcal{O}(1)$ & 80.77 \\
         & \textbf{EASE \citep{zhou2024expandable}} & $1\!-\!3\% \times T$ & $n+p \cdot T$ & $\mathcal{O}(T)$ & $\mathcal{O}(T)$ & 83.32 \\
         \midrule
         \multirow{3}{*}{First Session Adaptation} 
         & \textbf{ADAM~\citep{AdamAdapter}} & $1\!-\!3\%$ & $n+p$ & $\mathcal{O}(1)$ & $\mathcal{O}(1)$ & 80.02 \\
         & \textbf{RanPAC \citep{ranpac}} & $1\!-\!3\%$ & $n+p$ & $\mathcal{O}(1)$ & $\mathcal{O}(1)$ & 82.69 \\
         & \textbf{\method (Ours)} & $1\!-\!3\%$ & $n+p$ & $\mathcal{O}(1)$ & $\mathcal{O}(1)$ & 83.36 \\
         \bottomrule
    \end{tabular}
    }
    \label{tab:complexity_tab}
\end{table}

\section{Effect of Varying Task Sequence Orders}
\label{sec:TaskOrder}
This section evaluates the adaptability of our proposed method \mame{} and compares it with ADAM in CIL scenarios on the 5-Datasets benchmark. We specifically delve deeper into the experiments discussed in Sec.~4.3 of the main paper, where we analyze the influence of task order. on the impact of the task order. We provide more detailed experimental results considering multiple task orders. The task sequences include variations with SVHN, MNIST, CIFAR10, NotMNIST, and FashionMNIST. Results are reported in Tab.~\ref{tab:benchmark5datasets}. The orange cells corresponding to the first task, both ADAM and \mame{} have been trained on.

The results indicate the substantial influence of task sequence on model performance, highlighting the pivotal role of the initial task. \mame{} consistently surpasses ADAM in all sequences, with a maximum performance gain of +10.29\% observed in the SVHN \(\rightarrow\) MNIST \(\rightarrow\) CIFAR10 \(\rightarrow\) NotMNIST \(\rightarrow\) FashionMNIST sequence. This suggests that \mame{} effectively leverages unlabeled data from subsequent tasks to improve its adaptability.

\begin{table*}[h]
	\caption{Performance Comparison of ADAM and \mame{}          Across Varying Task Sequence Orders in Class-Incremental Learning. Top-1 accuracy is reported ($\uparrow$). Orange cells indicate the task that the model has been trained on. Model used in ViT-B/16 pre-trained on Imagenet-21K.
	}
        \label{tab:benchmark5datasets}
	\centering
		\resizebox{0.8\textwidth}{!}{%
		\begin{tabular}{@{}c|cccccc|c}
			\toprule
			\multicolumn{1}{c}{\multirow{1}{*}{Method}} & 
                &
			\multicolumn{1}{c}{SVHN}
			& \multicolumn{1}{c}{MNIST}
			& \multicolumn{1}{c}{CIFAR10}
			& \multicolumn{1}{c}{NotMNIST}
			& \multicolumn{1}{c}{FashionMNIST}
                & \multicolumn{1}{c}{Average} \\
                \midrule
            NCM & & 33.18&	83.78&	95.63&	68.85&	83.18&	72.92\\		
  \midrule
  \multirow{5}*{{ADAM~\citep{AdamAdapter}}}
            &SVHN & \cellcolor{orange!50}33.44 & 83.84 & 95.62 & 68.85 & 83.14 & 72.98 \\
            &MNIST & 92.44 &	\cellcolor{orange!50}93.95 &	72.77 &	83.00 &	40.22 &	76.48 \\
            &CIFAR10 & 97.24	& 67.97 &	\cellcolor{orange!50}83.63 &	31.33 &	84.04 &	72.84 \\
            &NotMNIST & 71.02 &	83.66 &	38.20 &	\cellcolor{orange!50}86.51 &	 95.89 &	75.06 \\
            &FashionMNIST & 84.19 &	31.52 &	85.69 &	95.96 &	\cellcolor{orange!50}69.93 &	73.46 \\
            \midrule
            \multirow{5}*{{Ours}}
            &SVHN&	\cellcolor{orange!50}95.13&	96.48&	67.33&	78.43&	78.96&	\textbf{83.27} \\
            &MNIST&	97.89&	\cellcolor{orange!50}76.15&	82.35&	83.09&	53.28&	78.55 \\
            &CIFAR10&	98.95&	70.37&	\cellcolor{orange!50}83.24&	35.03&	82.19&	73.96 \\
            &NotMNIST&	88.67&	83.82&	59.16&	\cellcolor{orange!50}92.76&	73.61&	\underline{79.60} \\
            &FashionMNIST&	91.02&	44.35&	84.63&	82.50&	\cellcolor{orange!50}77.78&	76.06 \\
            \bottomrule
		\end{tabular}
			}
\end{table*}

\section{Evaluation on CIFAR100-C with Level-5 Corruptions}
\label{sec:corruptions}
We compare our method, \name, with state-of-the-art methods on CIFAR100-C with Level-5 corruptions~\citep{hendrycks2019robustness}, as summarized in Tab.~\ref{tab:benchmark_cifar100}. The NCM performs poorly, yielding a 53.89\% average accuracy, due to its reliance on pre-trained representations. ADAM improves upon this with a 66.57\% average accuracy, courtesy of its initial task training.

In contrast, \name achieves the highest average accuracy of 68.00\%, outperforming ADAM by a significant margin of +5.13\% on CIFAR100-C. These results substantiate \name's superior adaptability and robustness to a variety of corruptions. 

\begin{table*}[t]
	\caption{\small Average and last performance comparison on seven datasets with {\bf ViT-B/16-IN21K} as the backbone on \emph{CIFAR100-C} with Level-5 corruptions~\citep{hendrycks2019benchmarking}. The best performance is shown in bold, and the second-best is underlined.}
 	\vspace{-3mm}
	\centering
		\resizebox{1.0\textwidth}{!}{%
		\begin{tabular}{l|ccc|cccc|cccc|c}
  
			\toprule
			\multicolumn{1}{l}{\multirow{2}{*}{Method}} & 
			\multicolumn{3}{c}{Noise} & 
                \multicolumn{4}{c}{Blur} &
			\multicolumn{4}{c}{Weather} &
                \multicolumn{1}{c}{Average} \\
        		& Gauss. & Shot & Impul. 
        		& Defoc. & Glass & Motion & Zoom 
                    & Snow & Frost & Fog & Brit. \\
		\midrule
        NCM	& 35.52&	39.05&	46.89&	59.78&	31.29&	56.99&	65.04&	64.41&	63.99&	50.13&	79.68&	53.89  \\
	ADAM~\citep{AdamAdapter}   & 37.52&	48.73&	65.03&	78.02&	39.80&	74.59&	79.27&	76.92&	73.34&	70.50&	88.59&	\underline{66.57}\\
        \midrule
	\name & 47.33&	51.45&	67.89&	78.52&	42.70&	75.00&	79.00&	75.31&	72.49&	69.20&	89.09 & \textbf{68.00} \\
			\bottomrule
		\end{tabular}
			}
   \label{tab:benchmark_cifar100}
\end{table*}

\section{Ablation Study: Augmentations, Iterations, and Batch Size}
\label{sec:ablations}
This section provides the additional ablation studies on the effect of augmentations $M$, iterations $N$, and batch size $B$. 

\begin{table}[H]
    \centering
    \renewcommand{\arraystretch}{1.2} % Improve row spacing
    \begin{minipage}{0.48\textwidth} % Text on the left
        \noindent \textbf{Effect of adaptation augmentations $M$. }  
        This section investigates the impact of varying the number of augmentations $M$ used during Phase II, as summarized in Tab.~\ref{tab:augment_summary}. Our method, \mame{}, demonstrates robust performance across both VTAB and Imagenet-R datasets, irrespective of the number of augmentations. However, increasing $M$ incurs computational costs during inference without a proportional performance gain. Thus, for efficiency, we recommend limiting augmentations to a small range \( M \in [1, \cdots, 8] \).
    \end{minipage}%
    \hfill
    \begin{minipage}{0.48\textwidth} % Table on the right
        \centering
        \caption{Summarized results for VTAB and Imagenet-R datasets based on the number of augmentations $M$. (Mean $\pm$ Std. Dev.)}
        \label{tab:augment_summary}
        \resizebox{0.95\textwidth}{!}{
        \begin{tabular}{c|cc}
            \toprule
            Augmentations (M) & VTAB$\uparrow$ & Imagenet-R$\uparrow$ \\
            \midrule
            4 & 90.14 $\pm$ 0.72 & \textbf{75.30 $\pm$ 0.87} \\
            8 &  \underline{90.55 $\pm$ 0.61} & 75.22 $\pm$ 0.91 \\
            16 & \textbf{90.67 $\pm$ 0.39} & 75.26 $\pm$ 0.85 \\
            24 & 90.52 $\pm$ 0.36 & \underline{75.23 $\pm$ 0.80} \\
            \bottomrule
        \end{tabular}
        }
    \end{minipage}
\end{table}

\begin{table}[H]
    \centering
    \renewcommand{\arraystretch}{1.2} % Improve row spacing

    \begin{minipage}{0.48\textwidth} % Text on the left
        \noindent \textbf{Effect of adaptation iterations $N$. }  
        Tab.~\ref{tab_iter2} summarizes the impact of increasing test-time iterations. Performance degrades as $N$ increases across both datasets. Specifically, VTAB accuracy drops from 91.28\% with one iteration to 80.78\% at four iterations, while Imagenet-R follows a similar trend, declining from 76.88\% to 69.13\%. These results indicate that excessive test-time adaptation may lead to overfitting or instability.
    \end{minipage}%
    \hfill
    \begin{minipage}{0.48\textwidth} % Table on the right
        \centering
        \caption{Impact of adaptation iterations $N$.}
        \label{tab_iter2}
        \resizebox{0.95\textwidth}{!}{
        \begin{tabular}{c|cc}
            \toprule
            Iterations & VTAB ($\bar{\mathcal{A}}$) $\uparrow$ & Imagenet-R ($\bar{\mathcal{A}}$)$\uparrow$ \\
            \midrule
            $N=1$ & \textbf{91.28} $\pm$ 0.64 & \textbf{76.88} $\pm$ 1.02 \\
            $N=2$ & \underline{87.34} $\pm$ 3.05 & \underline{74.77} $\pm$ 1.22 \\ 
            $N=3$ & 84.07 $\pm$ 3.53 & 70.14 $\pm$ 3.35 \\
            $N=4$ & 80.78 $\pm$ 5.45 & 69.13 $\pm$ 2.08 \\
            \bottomrule
        \end{tabular}
        }
    \end{minipage}
\end{table}

\begin{table}[H]
    \centering
    \renewcommand{\arraystretch}{1.2} % Improve row spacing

    \begin{minipage}{0.48\textwidth} % Text on the left
        \noindent \textbf{Effect of Batch-size $B$. }  
        Tab.~\ref{tab:batchsize} reports the impact of batch size $B$ on performance. For VTAB, accuracy stabilizes around 90\% for $B \geq 4$, with minimal gains from further increases. Conversely, Imagenet-R shows a decline in performance as $B$ increases, suggesting that a mix of low- and high-entropy samples may negatively impact the TTA process~\citep{niu2022efficient}.
    \end{minipage}%
    \hfill
    \begin{minipage}{0.48\textwidth} % Table on the right
        \centering
        \caption{Impact of batch size $B$ on VTAB and ImageNet-R performance in \mame{}. Results are reported as mean $\pm$ standard deviation.}
        \label{tab:batchsize}
        \resizebox{0.95\textwidth}{!}{
        \begin{tabular}{c|cc}
            \toprule
            Batch Size (B) & VTAB$\uparrow$ & Imagenet-R$\uparrow$ \\
            \midrule
            1 & 88.14 $\pm$ 1.71 & \textbf{79.11 $\pm$ 1.46} \\
            4 & 89.81 $\pm$ 2.62 & \underline{78.22 $\pm$ 2.54} \\
            8 & 88.91 $\pm$ 2.87 & 76.62 $\pm$ 1.93 \\
            16 & \textbf{91.28} $\pm$ 0.64 & 76.88 $\pm$ 1.02 \\
            32 & \underline{90.31 $\pm$ 0.09} & 69.12 $\pm$ 0.60 \\
            \bottomrule
        \end{tabular}
        }
    \end{minipage}
\end{table}

\begin{table}[h]
\centering
\begin{tabular}{l|cc|cc}
\toprule
\textbf{Method} & \multicolumn{2}{c|}{\textbf{CUB-Inc10}} & \multicolumn{2}{c}{\textbf{IN-R-Inc5}} \\
 & FWT & BWT & FWT & BWT \\
\midrule
ADAM~\citep{AdamAdapter}   & -0.47 & -4.64 & -1.68 & -11.96 \\
RanPAC~\citep{ranpac}   & -0.24 & -2.99 & -0.78 & -10.92 \\
\midrule
\rowcolor{lightgreen}
\textbf{PLASTIC (Ours)}  & \textbf{2.29} & \textbf{-1.92} & \textbf{-0.21} & \textbf{-10.65}  \\
\bottomrule
\end{tabular}
\caption{Comparison of forward transfer (FTW) and backward transfer (BWT) on CUB and ImageNet-R across different baselines.}
\label{tab:fwt_bwt_comparison}
\end{table}

\section{Data Efficiency}
This section presents the results of the ablation study conducted in Sec.4.3 of the main paper, which focuses on the data efficiency of our method \mame{}. Tab.~\ref{table:scarcedata} details the results presented in Fig.~4 in the main paper.

The study aims to examine how well \mame{} performs under varying degrees of data scarcity for training. We compare \mame{}'s performance against state-of-the-art methods like NCM and ADAM across different datasets including VTAB and Imagenet-R. 

\begin{table*}[t]
    \centering
    \caption{Performance comparison under varying data scarcity for training.}
    \vspace{-3mm}
    \begin{tabular}{c|cc|cc|cc}
        \toprule
        \multirow{2}{*}{\textbf{Training Data (\%)}} 
        & \multicolumn{2}{c}{\textbf{NCM}} 
        & \multicolumn{2}{c}{\textbf{ADAM}} 
        & \multicolumn{2}{c}{\textbf{\method (Ours)}} \\
        %\cline{2-7}
        & VTAB & Imagenet-R & VTAB & Imagenet-R & VTAB & Imagenet-R \\
        \midrule
        20\%  & \(87.52 \pm 0.91\) & \(56.27 \pm 0.45\) & \(87.49 \pm 0.91\) & \(56.75 \pm 0.40\) & \(88.00 \pm 0.03\) & \(57.39 \pm 0.29\) \\
        40\%  & \(88.84 \pm 0.76\) & \(59.41 \pm 1.42\) & \(88.78 \pm 0.66\) & \(63.34 \pm 1.51\) & \(89.57 \pm 0.88\) & \(65.66 \pm 1.15\) \\
        60\%  & \(89.51 \pm 0.69\) & \(60.82 \pm 1.10\) & \(89.56 \pm 0.71\) & \(68.67 \pm 1.08\) & \(90.29 \pm 0.14\) & \(70.89 \pm 1.09\) \\
        80\%  & \(89.43 \pm 0.90\) & \(61.18 \pm 1.32\) & \(89.61 \pm 0.75\) & \(72.05 \pm 0.82\) & \underline{\(90.38 \pm 0.24\)} & \underline{\(73.98 \pm 0.75\)} \\
        
        100\% & \(89.74 \pm 0.78\) & \(61.93 \pm 1.49\) & \(89.90 \pm 0.67\) & \(73.59 \pm 1.34\) & \textbf{90.54} \textbf{$\pm$} \textbf{0.64} & \textbf{75.24} \textbf{$\pm$} \textbf{0.91} \\
        \bottomrule
    \end{tabular}
    \label{table:scarcedata}
\end{table*}

\section{Inference cost}
\label{sec:inference}
\begin{table}[H]
    \centering
    \renewcommand{\arraystretch}{1.2} % Improve row spacing

    \begin{minipage}{0.48\textwidth} % Text on the left
        \noindent \textbf{Inference Cost Analysis.}  
        \method introduces additional computation through an optimization step before prediction. We analyze its inference cost by comparing the inference time of \mame{} with ADAM and NCM. Specifically, inference time in \method includes both TTA optimization and final prediction. Tab.~\ref{tab:inference_time} reports the average inference time (ms) over five runs on an A100 (48GB) GPU, using one iteration ($N\!=\!1$) and $M\!=\!8$ augmentations. 

        The computational overhead in \mame{} is a trade-off for improved plasticity and robustness. However, this cost is mitigated by its efficiency during training, where only Layer Normalization parameters are updated in the initial task.
    \end{minipage}%
    \hfill
    \begin{minipage}{0.48\textwidth} % Table on the right
        \centering
        \caption{Inference Time Comparison (ms). Results are averaged over 5 runs. GFLOPS/GMACs are evaluated at batch size $B\!=\!16$.}
        \label{tab:inference_time}
        \resizebox{0.95\textwidth}{!}{
        \begin{tabular}{c|c c c}
            \toprule
            Batch Size & \method & NCM & ADAM \\
            \midrule
            B=16 & 1820 & 22 & 27 \\
            \midrule
            GFLOPS & 85.45 & \textbf{33.72} & \underline{34.18} \\
            GMACs & 42.73 & \textbf{16.86} & \underline{17.09} \\
            \bottomrule
        \end{tabular}
        }
    \end{minipage}
\end{table}
\begin{table}[H]
    \centering
    \renewcommand{\arraystretch}{1.2} % Improve row spacing

    \begin{minipage}{0.48\textwidth} % Text on the left
        \noindent \textbf{Execution Time Analysis.}  
        Tab.~\ref{tab:execution_time} quantifies \mame{}'s inference time across varying batch sizes $B$ and adaptation iterations $N$, highlighting computational trade-offs. Notably, execution time scales more gradually with larger batch sizes.

        For instance, increasing from $B=1$ to $B=4$ at $N=1$ raises inference time by 280 ms (from 190 ms to 470 ms). However, when increasing from $B=8$ to $B=16$, execution time rises by only 770 ms (from 1000 ms to 1770 ms). This suggests that \mame{} achieves a form of computational efficiency with larger batch sizes, reducing execution time per unit increase in batch size.
    \end{minipage}%
    \hfill
    \begin{minipage}{0.48\textwidth} % Table on the right
        \centering
        \caption{Execution Time Comparison (ms) for varying batch size $B$ and adaptation iterations $N$.}
        \label{tab:execution_time}
        \resizebox{0.95\textwidth}{!}{
        \begin{tabular}{cc|cccc}
            \toprule
            &&\multicolumn{4}{c}{\bf Iterations} \\
            && \small N=1 & \small N=2  & \small  N=3 & \small N=4 \\
            \midrule
            \parbox[t]{2mm}{\multirow{4}{*}{\rotatebox[origin=c]{90}{\bf Batch-size}}} 
            &\small B=1  & \textbf{190} & \underline{250} & 320 & 340 \\
            &\small B=4  & 470 & 630 & 680 & 810 \\
            &\small B=8  & 1000 & 1110 & 1270 & 1410 \\
            &\small B=16 & 1770 & 2140 & 2480 & 2460 \\
            \bottomrule
        \end{tabular}
        }
    \end{minipage}
\end{table}

\section{Adapter}
\label{sec:adapter}
The Adapter module, a bottleneck structure extensively studied in prior works~\citep{chenadaptformer, adapterNLP}, enables fine-tuning of Vision Transformer (ViT) outputs. Composed of a down-projection \(W_\text{down} \in \mathbb{R}^{d \times r}\), a non-linear activation function, and an up-projection \(W_\text{up} \in \mathbb{R}^{r \times d}\), it facilitates dimensionality reduction and subsequent restoration. We adopt the AdaptMLP structure as introduced in AdaptFormer~\citep{chenadaptformer}, effectively replacing ViT's native MLP. Given the input \( \mathbf{x}_\ell \) to the MLP% layer
, the output can be expressed as:
\begin{equation} \label{eq:adapt-adapter}
    \text{MLP}(\mathbf{x}_\ell) + s \cdot \text{ReLU}(\mathbf{x}_\ell W_\text{down})W_\text{up},
\end{equation}
where \( s \) is an optional, learnable scaling parameter. As done in ADAM~\citep{AdamAdapter}, during the adaptation phase, only adapter parameters are optimized, while the pre-trained weights of ViT remain fixed. We employ a hidden dimension $r$ of 16, resulting in approximately 0.3 million tunable parameters, a figure substantially lower than the 86 million parameters in the standard ViT-B/16 model.

\noindent \textbf{Ablation Study on Adapter Sizes in \method.} Tab.~\ref{tab:param_comparison} presents an ablation study examining the effect of varying adapter sizes \(r\) on the average performance across multiple datasets such as VTAB, CIFAR100, and ImageNet-R. The reported trainable parameters also scale with the adapter size, giving us an insight into the computational cost involved.

The table reveals that increasing the adapter size does not substantially improve the model's performance. For instance, while moving from $r\!=\!16$ to $r\!=\!256$, the average accuracy on the VTAB dataset slightly decreases from 89.65\% to 88.65\%. A similar trend is observed in the CIFAR100 and ImageNet-R datasets, with only marginal performance improvements or even slight deteriorations. 

More importantly, the increase in adapter size comes at a substantial computational cost. The number of trainable parameters rises exponentially with the adapter size, going from 0.30 Million for $r\!=\!16$ to 4.73M for $r\!=\!256$. This not only increases the model's complexity but also makes it computationally expensive to train.

In summary, our analysis underscores the conclusion that while increasing the adapter size in \mame{} may seem like a plausible avenue for performance improvement, it does not offer significant gains. Instead, it introduces computational inefficiencies due to the escalated number of trainable parameters, thereby questioning the trade-off between performance and computational cost.

\begin{table}[h]
    \centering
    \caption{Comparison of \mame{} Average performance Across Different Adapter Sizes $r$ across datasets. Reported number of trainable parameters is in Million (M).}
    \vspace{-3mm}
    \begin{tabular}{c|cccc}
    \toprule
                & $r=16$ & $r=32$ & $r=128$ & $r=256$ \\
    \midrule
    Params (M)	& 0.30	& 0.60	& 2.37	& 4.73 \\
    \midrule
    VTAB       & \textbf{91.57} & \underline{89.18} & 88.83 & 88.65 \\
    CIFAR100   & \textbf{92.34} & \underline{90.14} & 89.25 & 89.04 \\
    ImageNet-R & \textbf{76.88} & 75.03 & \underline{75.39} & 74.99 \\
    \bottomrule
    \end{tabular}
    \label{tab:param_comparison}
\end{table}

\section{Baselines}
\label{sec:baselines}

We briefly outline the methods against which our approach is evaluated:

\begin{itemize}
    \item \textbf{Finetune}: Incrementally trains on new datasets, inducing catastrophic forgetting as a consequence.
    \item \textbf{Finetune Adapter~\citep{chenadaptformer}}: Retains pre-trained weights while optimizing an adapter module. Classifiers specific to the current dataset \( \mathbf{D}_t \) are fine-tuned, while those for previous classes are held constant. %(\( \mathbf{w}_i, i \in \mathcal{Y}_{t-1} \))
    \item \textbf{LwF~\citep{li2017learning}}: Employs knowledge distillation~\citep{hinton2015distilling} as a regularizer to mitigate forgetting, relying on the legacy model for soft target generation.
    \item \textbf{L2P~\citep{wang2022learning}}: A leading PTM-based CIL method that maintains a frozen pre-trained model while optimizing a prompt pool. It incorporates a 'key-value' pairing mechanism for prompt selection and leverages an auxiliary pre-trained model for prompt retrieval.
    \item \textbf{DualPrompt~\citep{wang2022dualprompt}}: An extension of L2P that utilizes two categories of prompts—general and expert—for enhanced performance. It also uses an additional pre-trained model for prompt retrieval.
    \item \textbf{ADAM~\citep{AdamAdapter}}: fine-tunes the adapters~\citep{chenadaptformer} only on the first task, among the sequences of all the tasks in a dataset, to adapt to the dataset at hand.
    \imad{\item \textbf{RanPAC~\citep{ranpac}}: uses random projection layers with pre-trained models to effectively prevent catastrophic forgetting without requiring rehearsal memory.}

    \imad{\item \textbf{FeCAM~\citep{fecam}}: presents an exemplar-free continual learning method that uses a Bayes classifier with anisotropic Mahalanobis distance to model heterogeneous class distributions.}
    
    \item \textbf{SLCA~\citep{SLCA}}: improves the classification layer by modeling the class-wise distributions and aligning the classification layers in a post-hoc fashion.
    \item \textbf{EASE~\citep{zhou2024expandable}}: improves the classification layer by modeling the class-wise distributions and aligning the classification layers in a post-hoc fashion.
    
\end{itemize}

\section{Datasets}
\label{sec:supp_dataset}
We describe the datasets utilized in our study, summarized in Tab.~\ref{tab:supp_dataset}. While CIFAR100, CUB200, and ImageNet-R are established CIL benchmarks~\citep{rebuffi2017icarl,wang2022dualprompt,zhou2021co}, ImageNet is ill-suited for PTM-based CIL evaluation due to data overlap~\citep{wang2022learning}. Consequently, we introduce four additional benchmarks characterized by non-overlap with ImageNet, substantial domain diversity, and large-scale, cross-domain instances.

\begin{itemize}
    \item \textbf{CIFAR100}~\citep{krizhevsky2009learning}: Comprises 100 classes, 60,000 images—50,000 for training and 10,000 for testing.
    \item \textbf{CUB200}~\citep{WahCUB2002011}: Focuses on fine-grained visual categorization, containing 11,788 bird images across 200 subcategories, with 9,430 for training and 2,358 for testing.
    \item \textbf{ImageNet-R}~\citep{hendrycks2021many}: Extended for CIL by~\citep{wang2022dualprompt}, includes various styles and hard instances, totaling 24,000 training and 6,000 testing instances.
    \item \textbf{ImageNet-A}~\citep{hendrycks2021natural}: Features real-world adversarially filtered images, with 5,981 training and 1,519 testing instances.
    \item \textbf{OmniBenchmark}~\citep{zhang2022benchmarking}: A diverse benchmark challenging PTM generalization, containing 89,697 training and 5,985 testing instances across multiple semantic realms.
    \item \textbf{VTAB}~\citep{zhai2019large}: Encompasses 19 tasks across three categories—Natural, Specialized, and Structured. We select five datasets to construct a cross-domain CIL setting. Similar to ADAM~\citep{AdamAdapter},  we select 5 to construct a cross-domain class-incremental learning setting, i.e., Resisc45, DTD, Pets, EuroSAT, and Flowers.
    \item \textbf{5-datasets}~\citep{ebrahimi2020adversarial}: a sequence of classification datasets including SVHN, CIFAR10, not-MNIST, Fashion-MNIST and, and MNIST.
\end{itemize}

\begin{table}[t]
	\caption{Introduction about benchmark datasets.
		ObjectNet, OmniBenchmark, and VTAB contain massive classes, and we sample a subset from them to construct the incremental learning task.}
	\label{tab:supp_dataset}
	\centering
	\resizebox{1.0\linewidth}{!}{
		\begin{tabular}{lcccccc}
			\toprule
			\textbf{Dataset} & \# \textbf{training}  & \# \textbf{testing}  & \# \textbf{Classes}& \textbf{Link} \\ 
            &\textbf{instances}&\textbf{instances}\\\midrule
			CIFAR100 & 50,000 & 10,000 & 100 & \href{https://www.cs.toronto.edu/~kriz/cifar.html}{\nolinkurl{Link}}\\
			CUB200 & 9,430 & 2,358 & 200 & \href{https://www.vision.caltech.edu/datasets/cub_200_2011/}{\nolinkurl{Link}}\\
			ImageNet-R & 24,000 & 6,000 & 200 & \href{https://github.com/hendrycks/imagenet-r}{\nolinkurl{Link}}\\
			\midrule
			ImageNet-A & 5,981 & 1,519 & 200 & \href{https://github.com/hendrycks/natural-adv-examples}{\nolinkurl{Link}}\\
			ObjectNet & 26,509 & 6,628 & 200 & \href{https://objectnet.dev/}{\nolinkurl{Link}}\\
			OmniBenchmark & 89,697 & 5,985 & 300 & \href{https://github.com/ZhangYuanhan-AI/OmniBenchmark}{\nolinkurl{Link}}\\
		VTAB & 1,796 & 8,619 & 50 & \href{https://google-research.github.io/task_adaptation/}{\nolinkurl{Link}}\\
            5-datasets & 180,869 & 506.906 & 50 & \href{https://github.com/facebookresearch/Adversarial-Continual-Learning}{\nolinkurl{Link}}\\
			\bottomrule
		\end{tabular}
		}
\end{table}

\end{document}